\documentclass{article}

\PassOptionsToPackage{numbers}{natbib}

\usepackage[preprint]{neurips_2022}

\usepackage[utf8]{inputenc} 
\usepackage[T1]{fontenc}    
\usepackage{hyperref}       
\usepackage{url}            
\usepackage{booktabs}       
\usepackage{amsfonts}       
\usepackage{nicefrac}       
\usepackage{microtype}      
\usepackage{xcolor}         

\usepackage{bbm}
\usepackage{amsmath}

\usepackage{amssymb}
\usepackage{algorithm}
\usepackage{algpseudocode}
\usepackage{graphicx}
\graphicspath{{./figures/}}
\usepackage{caption}
\usepackage{subcaption}
\usepackage{acronym}
\usepackage{enumitem,kantlipsum}

\author{%
  David S.~Hippocampus\thanks{Use footnote for providing further information
    about author (webpage, alternative address)---\emph{not} for acknowledging
    funding agencies.} \\
  Department of Computer Science\\
  Cranberry-Lemon University\\
  Pittsburgh, PA 15213 \\
  \texttt{hippo@cs.cranberry-lemon.edu} \\
}

\usepackage{amsthm}

\title{Analysis of Branch Specialization and its Application in Image Decomposition}

\author{Jonathan Brokman, Guy Gilboa } 

\begin{document}

\maketitle

\begin{abstract}
Branched neural networks have been used extensively for a variety of tasks. Branches are sub-parts of the model that perform independent processing followed by aggregation. It is known that this setting induces a phenomenon called Branch Specialization, where different branches become experts in different sub-tasks. Such observations were qualitative by nature. In this work, we present a methodological analysis of Branch Specialization. We explain the role of gradient descent in this phenomenon, both experimentally and mathematically.  We show that branched generative networks naturally decompose animal images to meaningful channels of fur, whiskers and spots and face images to channels such as different illumination components and face parts.


\end{abstract}

\section{Introduction}
The use of branching is a common practice in many neural network architectures. From ensemble learning in the 90's \cite{jacobs1991adaptive}, \cite{jordan1994hierarchical}, to grouped convolutional blocks of the classification era \cite{krizhevsky2012imagenet}, \cite{xie2017aggregated}, \cite{szegedy2015going}, to the sparsely gated mixture of experts of today  \cite{shazeer2017outrageously}, \cite{zoph2022designing}, \cite{fedus2021switch}.

A branched model propagates information to several processing units called branches, usually of identical architecture. The branches do not communicate with each other and produce outputs which are then combined and aggregated. 

In the case of ensembles, where each sub-network is trained separately - \cite{fort2019deep}  shown a typical disagreement between the sub-networks' predictions, even though performace of each sub-network on the task at hand (image classification) was similar. In their experiments they observed that two disagreeing sub-networks where attained at two disjoint basins of attraction of the loss landscape.

In the case of mixture of experts, which are trendy today, the branches are trained together alongside a routing function - which selects the relevant branch(es) for the selected input and task. Different normalization and routing techniques are used to ensure that each branch learns a specific specializations. Nevertheless it has been observed early on, for example in AlexNet \cite{krizhevsky2012imagenet}, that without any type of regularization or routing,  branched CNN sub-modules produce specialized filters, where each branch specializes in distinct image features. In this work we would like to address this phenomenon, characterize the specialization effect in branching and provide preliminary mathematical insights.

We restrict ourselves to a setting of cumulative aggregation as follows.
Let $X=\{x_1,..\,,x_N\}$, $Y=\{y_1,..\,,y_N\}$ be sets of $N$ inputs and corresponding desired outputs (labels or target images). 
We have $M$ branches of sub neural-networks, identical in their architecture. 
\begin{figure}[htb]
  \centering
  \includegraphics[width=70mm]{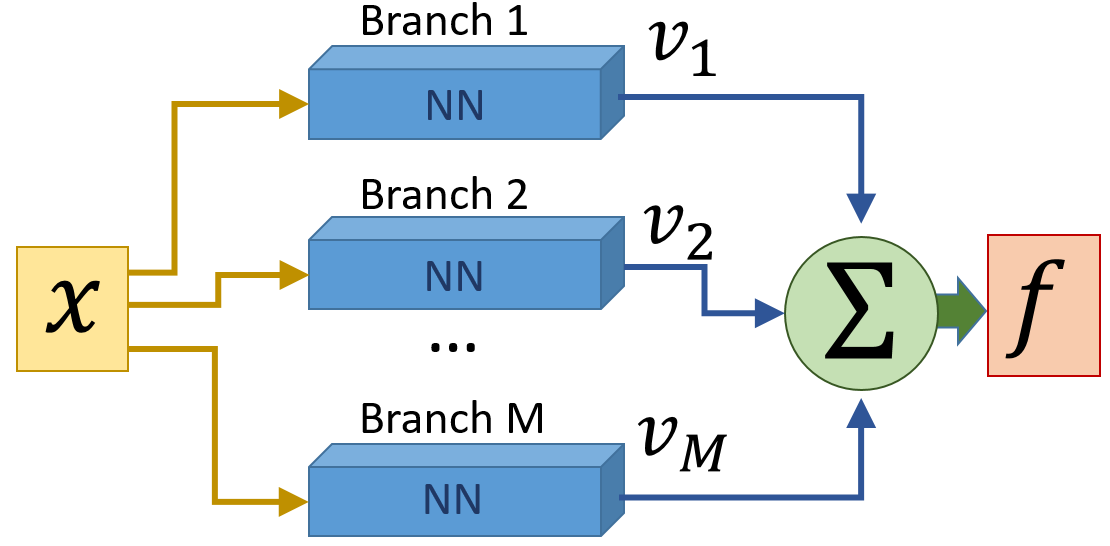}
  \caption{We examine a general branched neural network structure as illustrated above. Each branch consists of a sub neural network (blue). The response of all branches is summed to produce the global network output $f$. All sub neural networks have exactly the same architecture. The only difference is their different random initialization at the training phase.  }
  \label{fig:branch_diagram}
\end{figure}

For an input $x$ the output of each branch is $v_k(x)$ and the combined network output is  
\begin{equation}
\label{eq:f} 
    f_{\theta}(x) = \sum_{k=1}^M v_k(x),
\end{equation}
where $\theta$ is the set of model parameters. See Fig. \ref{fig:branch_diagram} for an illustration.
From associativity and cummutativity of summation the role of each branch is identical, in principle. The only difference is the different random parameter initialization of each sub-network, which occurs at the beginning of the training process. 
Nevertheless, this seemingly minor difference has a remarkable effect on  the role of the branch at inference. Through the gradient descent process, the branches spontaneously tend to  specialize in certain tasks of the global network goal, some become silent or ``null-networks'' which produce essentially zero output for any input $x$. Moreover, the correlation between the branches output is low and the tendency is not to share the same task by several branches. We show that this phenomenon is fundamental to branched architecture and happens from the smallest single scalar perceptron element, through standard CNN classifiers, to complex generative networks. In the generative case, we obtain an unsupervised decomposition of an image into various meaningful channels like  mouth and eyes, lightings of specular and diffusive components, and different fur textures and color patterns in animals. We sketch preliminary possible applications that such decompositions might yield.

\section{Previous work}
Throughout the last decade, Branch Specialization was repeatedly observed in various deep learning architectures and tasks, for instance \cite{krizhevsky2012imagenet}, \cite{shazeer2017outrageously}, \cite{zoph2022designing},  \cite{voss2021branch}.

Splitting networks to independent sub-parts dates back to \cite{jacobs1991adaptive}, \cite{jordan1994hierarchical}. It was popularized for CNNs by AlexNet \cite{krizhevsky2012imagenet}, which introduced grouped convolutions. This trend continued to re-appear in classification models \cite{szegedy2015going}, \cite{xie2017aggregated}.


Conditional computation, first suggested in \cite{bengio2013deep}, \cite{davis2013low}  splits the model to sub-parts too: A gating mechanism selects sub-parts to be used, conditioned on the input. This is efficient, since only a these parts are used for inference. Today its used both for NLP and for vision tasks, e.g. \cite{kirsch2018modular}, \cite{zoph2022designing},  \cite{shazeer2017outrageously}, \cite{fedus2021switch}  \cite{vaswani2017attention},  \cite{rosenbaum2017routing}, \cite{gregory2021hydranet}, \cite{cai2021dynamic}, \cite{xu2021meta}, \cite{ren2022beyond}.

Recently \cite{voss2021branch} have shown how dimensionality reduction techniques cluster convolutional kernels of branched classifiers by their branch. They termed this phenomenon \emph{Branch Specialization}, a term which we adopt.


In this work we systematically analyze branched models, and the naturally-occurring interactions between their branches. We analyze how branching affects gradient distribution, and experiment with several tasks of various degrees of complexity.

\section{Notations and Preliminaries \label{sec:setting}}
Denote the data domain $X$, a sample $x \in X$, a loss function $L(x)$, and a parameter-dependent function model $f_{\theta}:X\rightarrow\mathbb{R}^C$, where $\theta$ denotes the parameters and $C$ is the output dimension. We  assume the model is branched with $M$ branches as in Eq. \eqref{eq:f}. Branch $k$ is $v_k:X\rightarrow\mathbb{R}^C$, and depends on parameters $\theta_k \subset \theta$. Each branch is a neural-network, where different branches have identical architecture and differ by their parameters.

A reminder of common losses and their derivatives: For $L^2(x)=\frac{1}{2}||y - f_{\theta}(x)||_2^2$, we have
\begin{equation}\label{eq:collaborative_L2}
\frac{d L^2}{d f_{\theta}(x)}=y - f_{\theta}(x).
\end{equation}
For $L_{cross\, entropy}(x) = \sum_c p(x, c) log(s(f_{\theta}(x, c)))$, where $s$ is the softmax operator we get
 \begin{equation}\label{eq:cross_entropy_gradient}
 \frac{d L_{cross\, entropy}}{d f_{\theta}(x)} = p(x) - s(f_{\theta}(x)).  
 \end{equation}

\section{Main Observations and Toy Examples}


We summarize below the main observations on branching that are supported by our experiments (detailed hereafter). 


\begin{enumerate}
    \item Each network tends to specialize in certain aspects of the global problem. We refer to that as region of specialization (ROS). In classification it can be certain features belonging to certain classes. In image synthesis and generation - certain image characteristic (structural or semantic).
    \item The networks tend not to share ROS's. That is, in each ROS very few networks  are active (and often only a single one is active) .
    \item In order not to interfere in solving the global task, each network is silent (inactive) in most ROS's. 
    \item There may be completely silent networks, which are null (close to zero response) for any input data. This typically happens when
    the global task can be solved well by less than the total number of branches of the network.
    \item The response of each branch is lowly correlated to the response of other branches.
    \item The full network learning capacity grows with the number of branches $M$, as can be expected. However, the number of active branches may be almost constant for different $M$, if $M$ is large enough.
    \item The specialization process happens naturally through gradient descent optimization. We 
\begin{itemize}
\item 
\end{itemize}
provide theoretical support for that by analyzing the Hessian matrix.
    \item The properties above happen at all levels of complexity, from the neural level to very large, highly parameterized networks, in each branch.
\end{enumerate}




\subsection{Toy Examples}
\begin{figure}[htb]
  \centering
  \includegraphics[width=60mm]{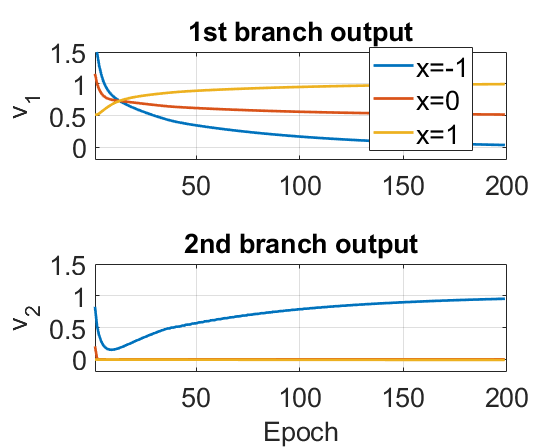}
  \includegraphics[width=60mm]{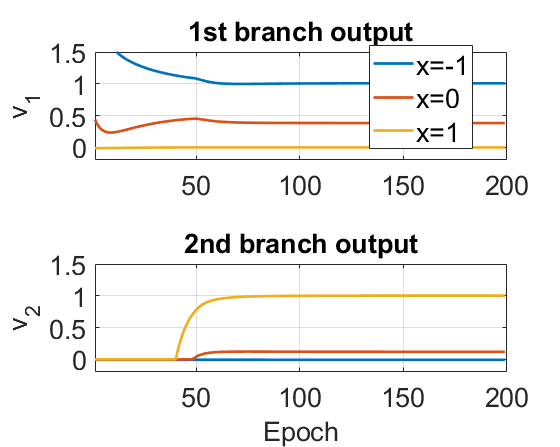}
  \caption{Dynamics of Toy Example 1 with 2 branches ($M=2$). Two cases are shown, left - a more common case of specialization and inactivity of the other branch in each region. On the right we see  two more rare phenomena: first for $x=1$ (yellow) both branches are inactive until the second branch spontaneously ``takes responsibility'' for this task. Second, sharing the task for $x=0$, where both branches are active.}
  \label{fig:toy1_dynamics}
\end{figure}

\begin{figure}[htb]
  \centering
  \includegraphics[height=30mm]{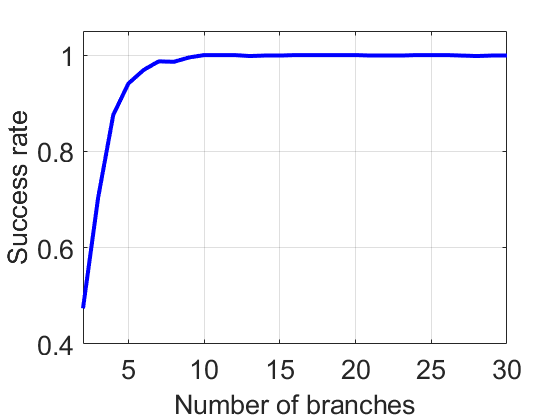}
    \includegraphics[height=30mm]{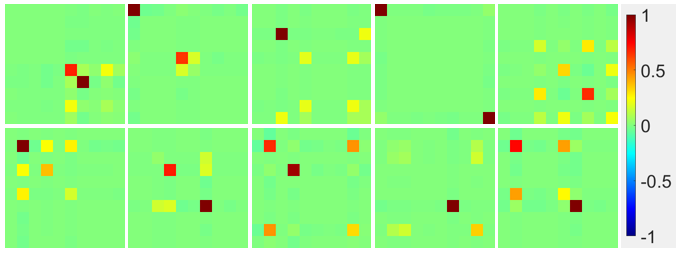}
  \caption{Toy 1. Left - success rate as a function of the number of branches in the network (1000 trials are performed for each branch configuration). Learning capacity grows with $M$ (although the number of active branches is similar, see below). Right - 10 examples of covariance matrices for 10 branches. The output of the different branches is loosely correlated.}
  \label{fig:toy1_success_rate_cov}
\end{figure}

We present here two very simple one dimensional toy examples. In both cases each branch $k$ is a neural component of the form:
\begin{equation}
\label{eq:toy_neuron}
    v_k(x) = \sigma(w_k\cdot x + b_k),
\end{equation}
where $\sigma$ is a nonlinear activation function (we use Leaky ReLU), $w_k$ is a scalar weight parameter and $b_k$ is a scalar bias parameter. Our combined network consists of $M$ such branches, where the network output $f_\theta$ is defined in \eqref{eq:f}.
The training set is composed of $N=4$ examples 
with an input set $X = \{x_1,.. \,\, x_N\}$ and a corresponding desired output set $Y = \{y_1,.. \,\, y_N\}$. The first toy example $Toy1$ consists of the training set  $X=\{-1, 0, 0, 1 \}$, $Y=\{1, 0.5, 0.5, 1 \}$. It is somewhat similar to a one-dimensional XOR problem, often used as toy data that a linear model cannot solve \cite{brutzkus2019larger}. To approximate a classification problem we have a balanced set of both classes $1$ and $0.5$. Note that we have not chosen in $Y$ a target value of $0$ since this value has a special role of null contribution in a sum, which we want to examine. The second toy problem $Toy2$ models a regression problem where $X = \{-1,0,1,2\}$ and $Y = \{1, 0.25, 0.5, 0.75\}$. A square $L^2$ loss is used $L=\frac{1}{2}\|f-y\|_2^2$.
Training is done using gradient descent. 
In Fig. \ref{fig:toy1_dynamics} we show two examples of the dynamics along the training for $Toy1$. In the caption some typical phenomena are explained. 
For this task we define success of the training if the loss is close to zero ($\sum_{j=1}^N(f(x_j)-y_j)^2 < \delta$, we chose $\delta=0.0001$). Thus we can measure success rate. This is shown in Fig. \ref{fig:toy1_success_rate_cov} (left). For a network configuration, with fixed $M$ (in the range 2 to 30), we train the network 1000 times  with different random initialization. 
It is clearly seen that as the number of branches increases - the optimization task is easier (success rate increases monotonically) and  around $100\%$ success rate is reached for 10 branches or more. 

The output of the branches is also loosely correlated.
We define the response matrix (of size $M \times N$) for all training set and all branches by $F$, where a matrix element at row $j$ and column $i$ is
$ F_{i,j}=v_i(x_j)$. The covariance matrix of the branches response is 
$ Cov =  F \cdot (F)^T$, where $T$ denotes  transpose. Examples for $M=10$ are shown in Fig. \ref{fig:toy1_success_rate_cov} (right), showing generally low covariance.

\begin{figure}[htb]
  \centering
  \includegraphics[height=30mm]{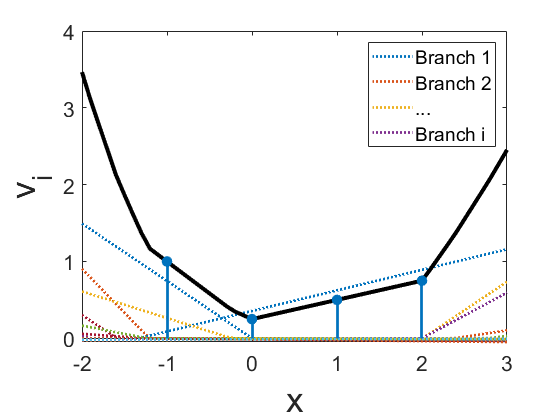}
    \includegraphics[height=30mm]{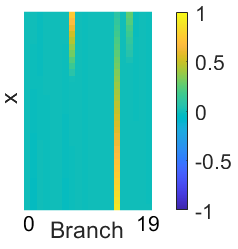}
    \includegraphics[height=30mm]{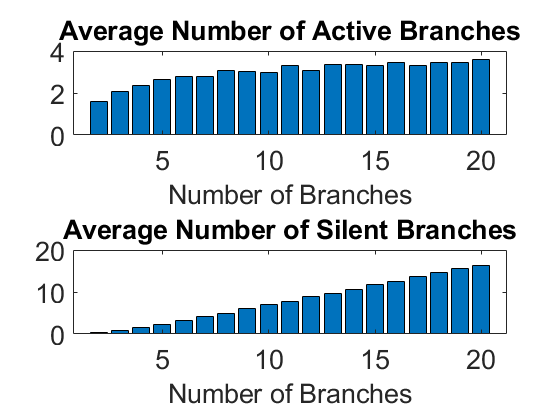}
  \caption{Toy 2. Left and middle - an example of the response of 20 branches. Right - average number of active branches for different $M$. We define an active branch as one where the norm of its response is at least $10\%$ of the maximal branch response norm. Silent branches are the complement of active. On the left we have an example of 3 active branches. The number of active branches is almost constant,  such that silent branches grow linearly with $M$. }
  \label{fig:toy2_response}
\end{figure}

Let us examine the gradient descent process. The gradient for some weight parameter $w_i$ for a certain data element $x_j$ is 
\begin{equation}
    \nabla_{w_i} L(x_j) = (f(x_j)-y_j)\cdot(\sigma'(x_j w_i + b_i)x_j,
\end{equation}
where $\sigma'(q)=d \sigma(q)/dq$. We can write it as,
\begin{equation}
    \label{eq:toy_grad_L}
    \nabla_{w_i} L(x_j) = D_{collab}\cdot D_{dist},
\end{equation}
where $D_{collab}=(f(x_j)-y_j)$ is the collaborative part and $D_{dist}=(\sigma'(x_j w_i + b_i)x_j$ is the distributive part. The collaborative part ensures that whenever the task is solved for this specific data element by the entire network, this element will not affect the parameters of the network. The distributive part becomes relevant when the collaborative part is not negligible and attempts to reduce the loss locally for the specific branch. 
We will later see that this characteristics can be generalized and tend to induce specialization.
Some examples and statistics of the two Toy examples are shown and explained in Figs. \ref{fig:toy1_dynamics}, \ref{fig:toy1_success_rate_cov} and \ref{fig:toy2_response}.

\section{Loss Gradient and Hessian of Branched Models\label{sec:mathematical}}

In the setting of Sec. \ref{sec:setting}, $f_{\theta}(x) \in \mathbb{R}^C$ is optimized w.r.t. a loss $L(x)$ on data $x \in X$. We analyze gradient descent (GD) optimization as the following nonlinear dynamical system,
\begin{equation} \label{eq:dynamical_system}
\dot{\theta}(t) = -\nabla_{\theta(t)}L(x),
\end{equation}
where $t$ is the time variable, and $\dot{\theta}=\frac{\partial \theta}{\partial  t}$.  The trained model is attained at an equilibrium point, $\dot{\theta}=0$. For stochastic GD (SGD) this translates to $\nabla_{\theta(t)}L(x) \approx 0, \,\forall x \in X$. The fixed point stability is characterized by the Jacobian of $\dot{\theta}$, in our case the Hessian matrix of $L(x)$ w.r.t $\theta$.  This type of analysis is common, e.g. \cite{li2020hessian}, \cite{sagun2016eigenvalues}, \cite{sagun2017empirical}, \cite{arora2018convergence}, \cite{du2018gradient}.

\subsection{Gradient}
Denote $\nabla_{\theta_k} v_k(x),\,\nabla_{\theta_k} f(x) \in \mathbb{R}^{C \times |\theta_k|}$ Jacobian matrices w.r.t. $\theta_k$. By the chain rule

\begin{equation}
\nabla_{\theta_k} L(x) = \left( \nabla_{\theta_k} f_{\theta}(x) \right) ^T \cdot \left(  \frac{d L}{d f_{\theta}(x)} \right),
\end{equation}

plugging Eq. \eqref{eq:f} we have by linearity

\begin{equation}
\nabla_{\theta_k} L(x) = \left(   \sum_{l=1}^M \nabla_{\theta_k} v_l(x) \right) ^T \cdot \left(  \frac{d L}{d f_{\theta}(x)} \right),
\end{equation}

and since $\theta_k \cap \theta_l = \emptyset \forall k \neq l$ we have

\begin{equation} \label{eq:collab_distrib}
\nabla_{\theta_k} L(x) = \underbrace{\left( \nabla_{\theta_k} v_k(x) \right) ^T}_\text{distributive} \cdot \underbrace{\left(  \frac{d L}{d f_{\theta}(x)} \right)}_\text{collaborative},
\end{equation}
or equivalently, for a single parameter $w_k \in \theta_k$,
\begin{equation} \label{eq:collab_distrib, scalar}
\frac{\partial L(x)}{\partial w_k} = \underbrace{\left(  \frac{ \partial v_k(x)}{\partial w_k}\right) ^T}_\text{distributive}  \cdot  \underbrace{\left( \frac{d L}{d f_{\theta}(x)}\right)}_\text{collaborative}.
\end{equation}

Gradient is factorized to distributive and collaborative parts, affecting a branch or the whole model.

\subsection{Hessian}
Let $w_k \in \theta_k,\,w_l \in \theta_l$. Using Eq. \eqref{eq:collab_distrib, scalar} we get
\begin{equation}
\frac{\partial^2 L(x)}{\partial w_k \partial w_l} = \frac{\partial}{\partial w_l} [\left( \frac{ \partial v_k(x)}{\partial w_k} \right) ^T \cdot \left( \frac{d L}{d f_{\theta}(x)}\right)].
\end{equation}
Case 1: If $\theta_l=\theta_k$ (but $w_k$ might be different than $w_l$) we get
\begin{equation}
\frac{\partial^2 L(x)}{\partial w_k \partial w_l} = \left( \frac{\partial}{\partial w_l} [\frac{ \partial v_k(x)}{\partial w_k} ]\right) ^T \cdot \left( \frac{d L}{d f_{\theta}(x)}\right) + \left( \frac{ \partial v_k(x)}{\partial w_k} \right) ^T \cdot \left( \frac{\partial}{\partial w_l} [\frac{d L}{d f_{\theta}(x)}]\right).
\end{equation}
Case 2: 
\begin{equation}\label{eq:mixed_jacobian_entries}
\theta_l \neq \theta_k \Rightarrow \frac{\partial^2 L(x)}{\partial w_k \partial w_l} =  \underbrace{\left(  \frac{ \partial v_k(x)}{\partial w_k}\right) ^T}_\text{distributive} \cdot \left( \frac{\partial}{\partial w_l} [\underbrace{ \frac{d L}{d f_{\theta}(x)}}_\text{collaborative}]\right).
\end{equation}
Note again the distributive and collaborative terms. When the distributive part of the gradient is zero, case 2 is zero, i.e. the Hessian becomes block diagonal, where a block corresponds to a branch.


\section{Image classification}
Here we train our neural networks on the CIFAR-10  dataset \cite{cifar10}. For this task, each branch $v_k$ is a "slimmed down" version of ResNet18 \cite{he2016deep}  as implemented at \cite{git_resnet} with $|\theta_k|=0.08 \cdot 10^6$ parameters. We sum $M=16$ branches,  thus $|\theta|=M\times |\theta_k|=1.3 \cdot 10^6$. For reference - ResNet18 uses about $11.2 \times 10^6$ parameters.
Optimization is done on the cross-entropy loss, between ground truth and learnt probability density functions (PDF) $p(x), q_{\theta}(x)$.
%
Summing an ensemble of logit vectors results with a valid logits vector, in contrast to PDF summation. Hence we apply softmax on $f_{\theta}(x)$, which makes summation of $\{v_k(x)\}_k$ valid.

For ease of analysis, we clamp $f_{\theta}(x)$ to $[-1, 1] \subset \mathbb{R}$. This restricts the logits from diverging to uncontrollably large values, which are hard to predict and analyze. Thus $p(x)$ is not a one-hot vector, but a softmax on the class-indicating $\pm1$ values.

\begin{figure}[htb]
  \centering
  \includegraphics[width=0.17\textwidth]{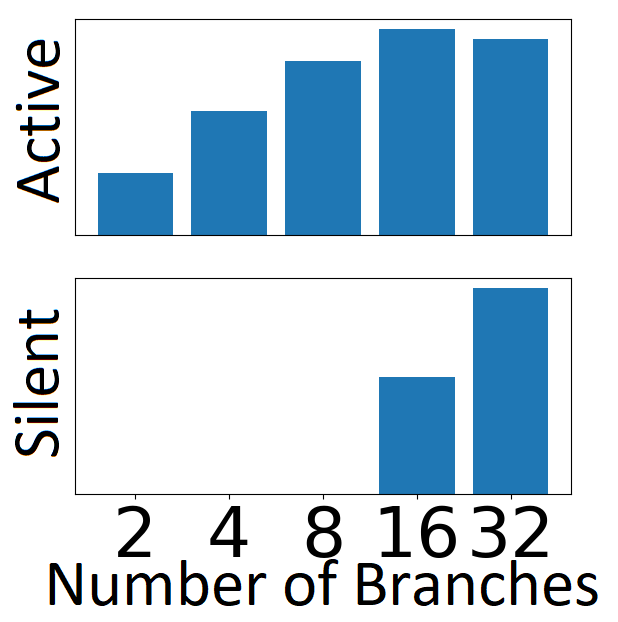}
  \includegraphics[width=0.245\textwidth]{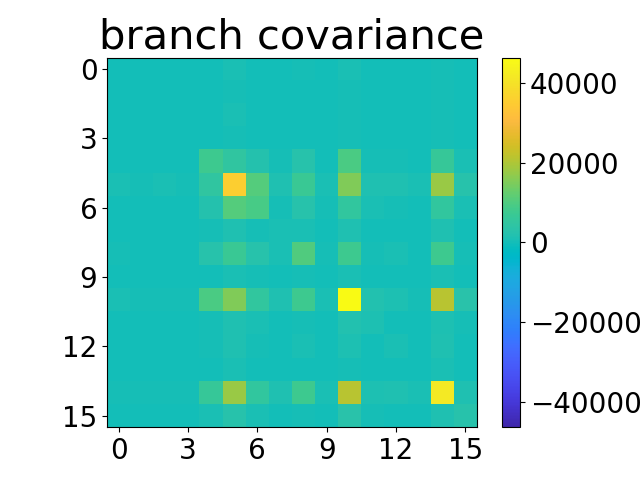}
  \includegraphics[width=0.245\textwidth]{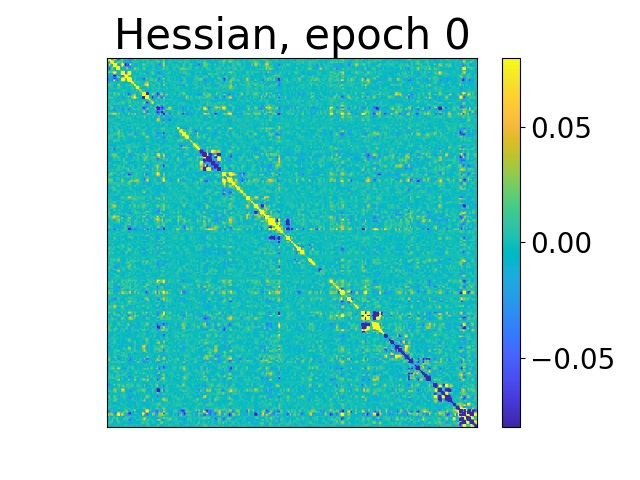}
  \includegraphics[width=0.245\textwidth]{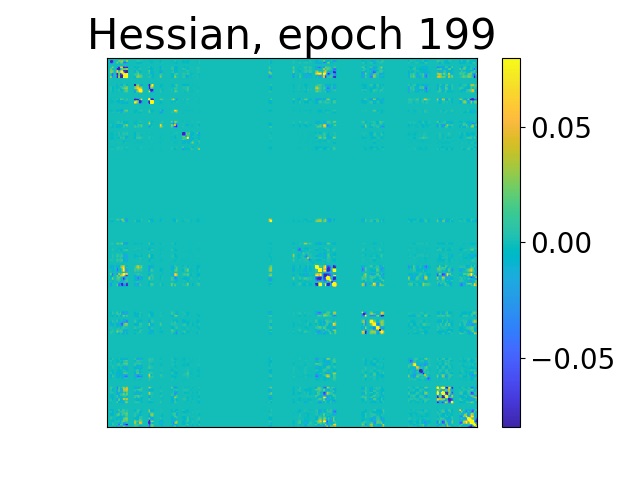}
  
  
  \caption{Classification. From left to right: 1) Logarithmic No. of Active and Silent Branches as function of $2^M$ where active branches are defined as in Fig. \ref{fig:toy2_response}.   2) Inter-branch covariance for $M=16$. 3-4) Hessian  w.r.t. parameters from the first layer of all 16 branches (at initialization and at training convergence). An almost block-diagonal Hessian is obtained, with 16 blocks corresponding to the 16 branches.}
  \label{fig: cifar correlation}
\end{figure}

\begin{figure}[htb]
  \centering
  \includegraphics[width=0.9\textwidth]{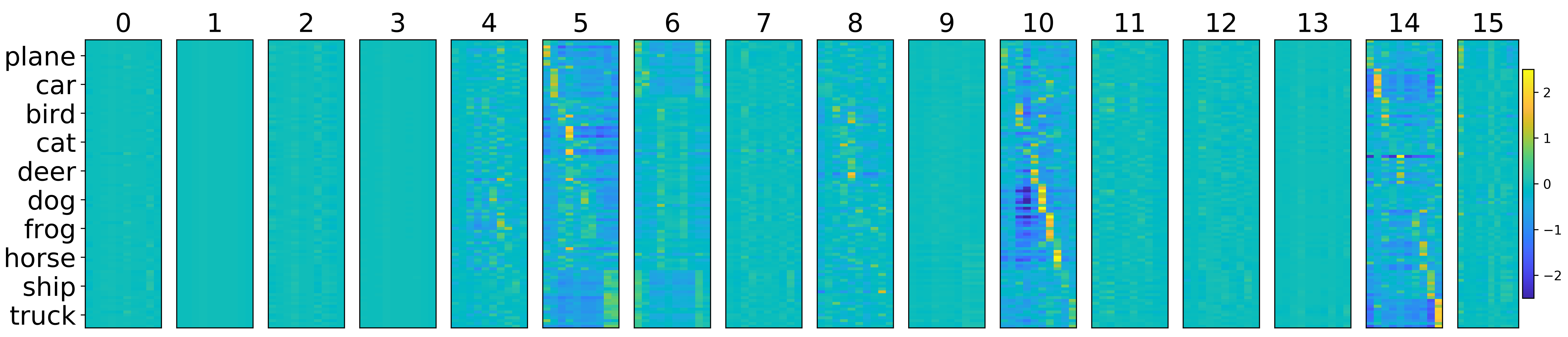}
  \caption{Classification, output by branch: Before summation, each branch produces its contribution to the classification inference. Branches differ by the amount of contribution - they range from highly important contributors to being completely turned off (Silent branches). The branches also differ by the classes they specialize in (but have also intra-class specialization).}
  \label{fig: cifar out by branch}
\end{figure}

\begin{figure}[htb]
  \centering

  \includegraphics[width=0.8\textwidth]{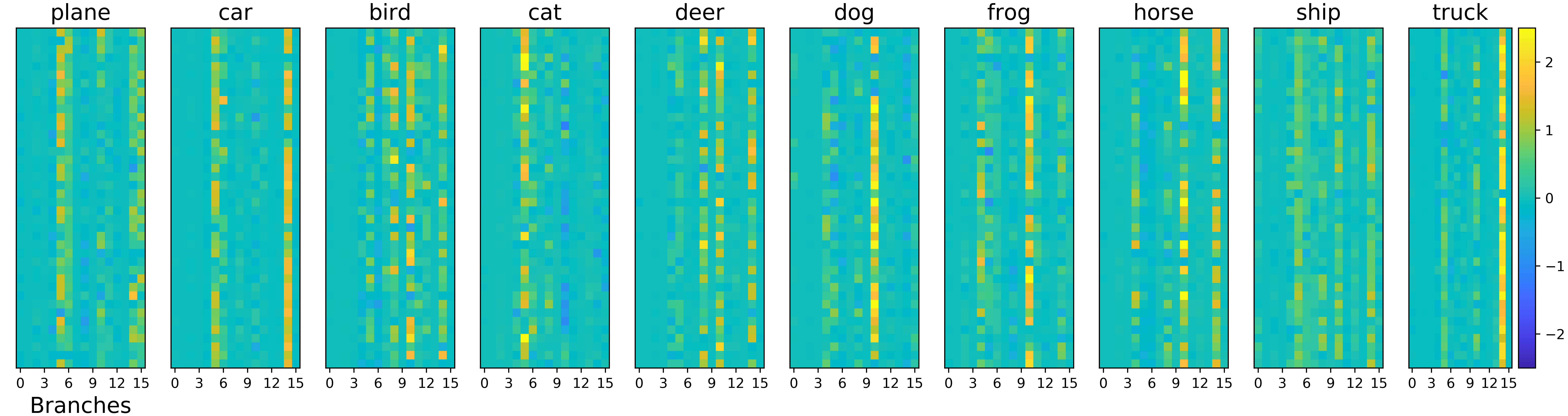}
  \caption{From each class, 20 samples were randomly selected, and the confidence of each branch in correctly classifying each sample is shown.  Larger positive values (yellow) indicate greater true confidence, and negative values (blue) - false confidence.  Processing is distributed differently between branches, depending on the input samples. We obtain both inter-class and intra-class specialization.}
  \label{fig: cifar selected class}
\end{figure}

\begin{figure}[htb]
  \centering
  \includegraphics[width=\textwidth]{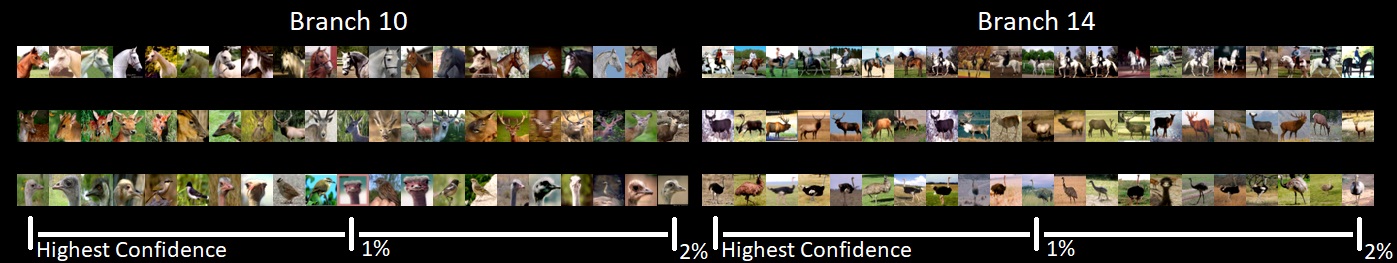}
  \caption{Intra-class specialization. We examine the two most informative branches - 10 and 14 (see Fig. \ref{fig: cifar out by branch}) on classes Horse, Deer and Bird. These classes have their classification inference distributed between several branches (see Fig. \ref{fig: cifar selected class}). For each branch-class pair , the confidence of all test samples was assessed, then samples with top $2\%$ confidence where selected for this figure. Samples are sorted by their confidence from left (highest confidence) to right. This exposes that branch 10 mainly specializes in classifying head close-up shots of horses, deer and large birds, while branch 14 specializes in whole body shots within these classes. Specifically - branch 14 seems to be more confident about horses with their equestrians.}
  \label{fig: branch look}
\end{figure}

Following training, Branch Specialization indeed occurs as can be observed in Fig. \ref{fig: cifar out by branch}. As in the toy examples - some branches  become experts in specific classes, while others are "turned off"  and do not contribute to classification. In Fig. \ref{fig: cifar correlation} we see that the covariance of branches is also consistent with the toy example. 
As can be seen in Fig. \ref{fig: cifar selected class} - different branches are confident about different cases. 
In Fig. \ref{fig: branch look} it is shown that branches specialize in certain characteristics within each class, here Branch 10 specialize in closer animals and Branch 14 in further ones.

\section{Class Transfer GAN}
So far we saw that different branches solve the task at hand from different aspects. Here we show how specialization is manifested in a generative model.

We adapt StarGAN-v2 \cite{choi2020stargan}, and train our neural network on the same datasets AFHQ  \cite{git_stargan}, and Celeb-a-HQ \cite{liu2015faceattributes}. While \cite{choi2020stargan} was designed for diverse image style-transfer, we restrict our discussion to non-diverse style transfer, or class transfer. Following Eq. \eqref{eq:f}, $f_{\theta}$ is now a Generator network, where each branch $v_k$ is a "slim version" of StarGAN-v2's Generator, using $M=9$ branches.


\begin{figure}[htb]
  \centering
  \includegraphics[width=18mm]{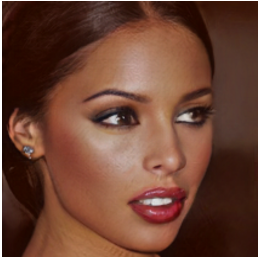}
  \includegraphics[width=18mm]{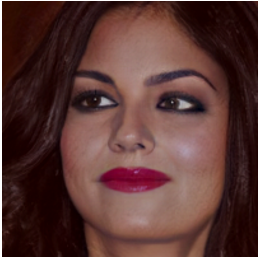}
  \includegraphics[width=18mm]{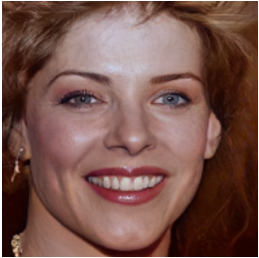}
  \includegraphics[width=18mm]{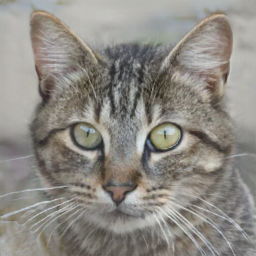}
  \includegraphics[width=18mm]{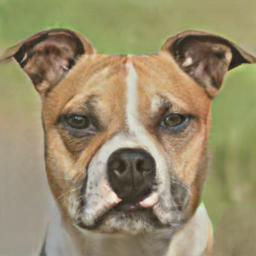}
  \includegraphics[width=18mm]{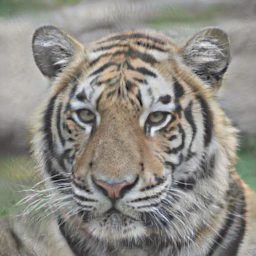}
  \caption{Celeb-a-HQ and AFHQ re-generated images: High quality and input fidelity are attained.}
  
  \label{fig:gan_generated}
\end{figure}

\begin{figure}[htb]
  \centering
  \includegraphics[width=40mm]{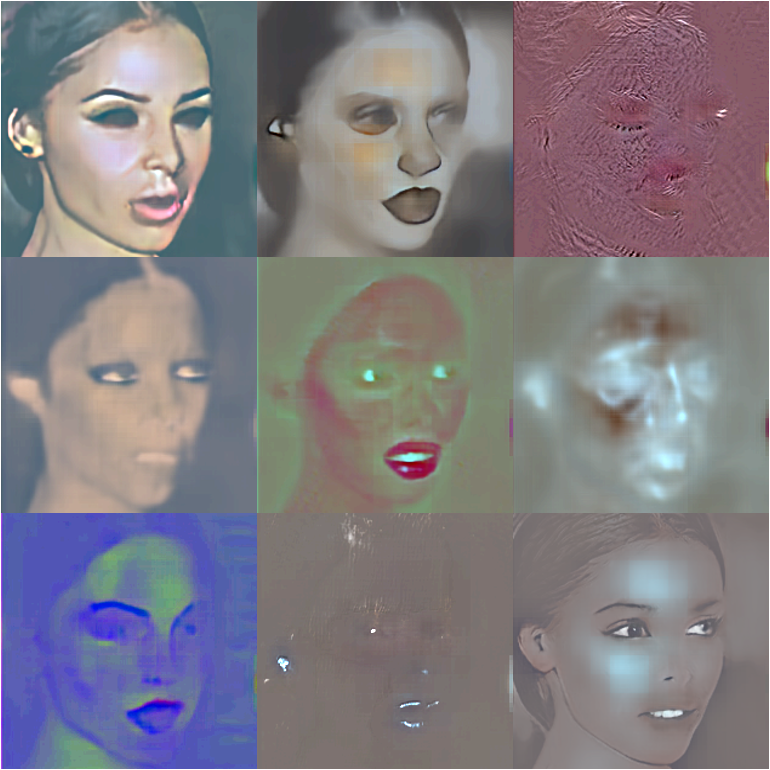}
  \includegraphics[width=40mm]{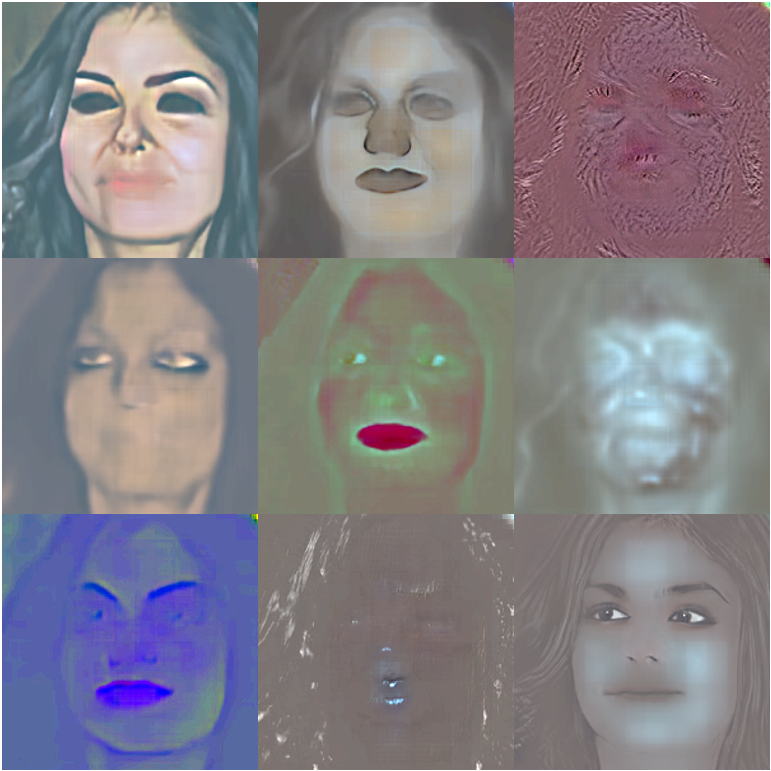}
  \includegraphics[width=40mm]{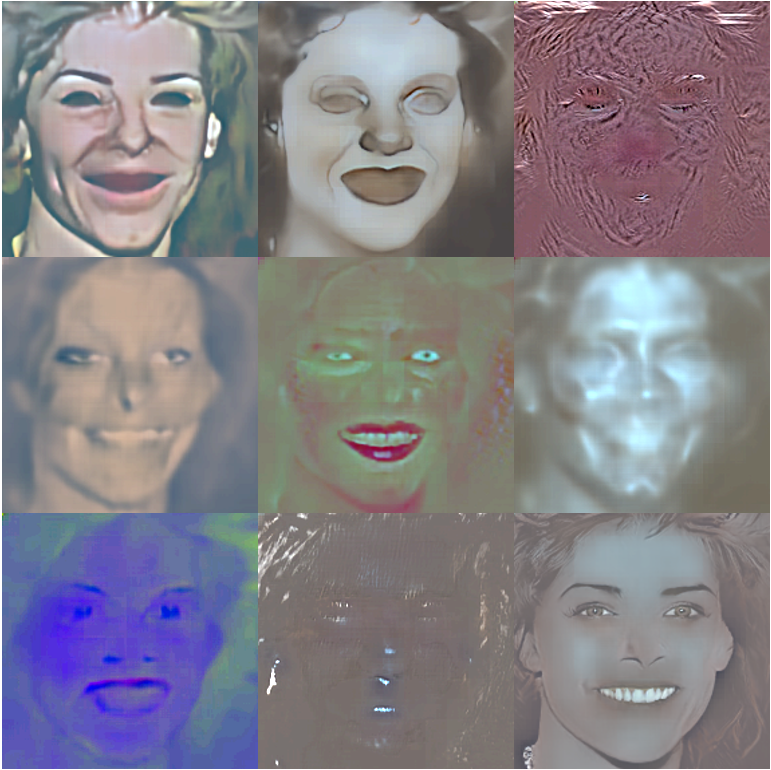}
  \includegraphics[width=40mm]{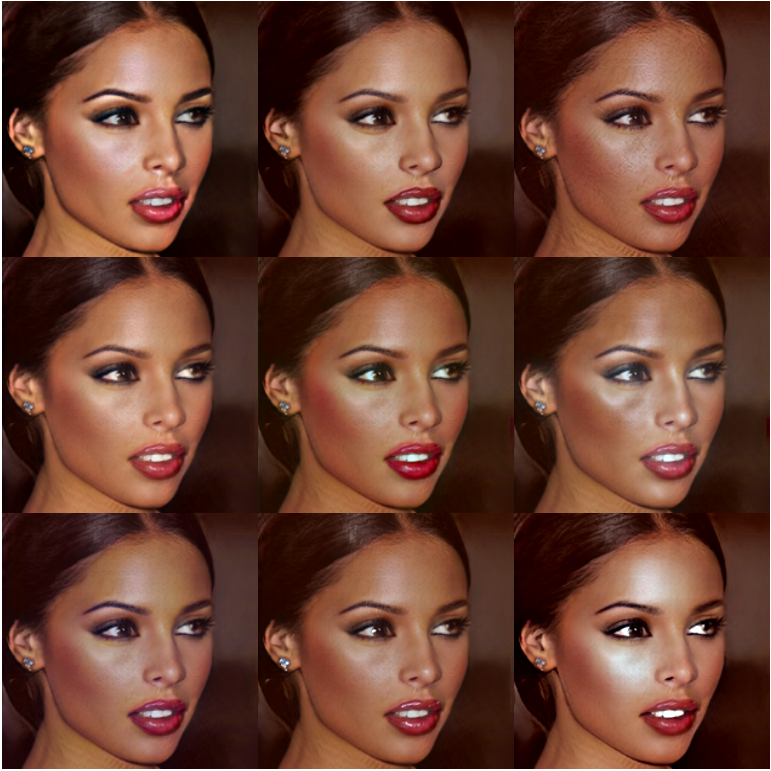}    
  \includegraphics[width=40mm]{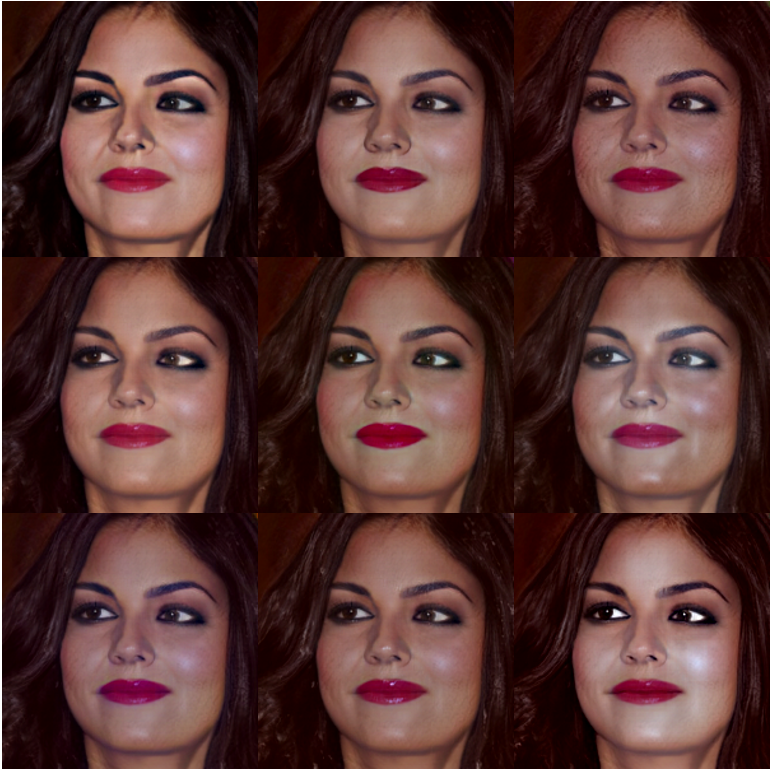}    
  \includegraphics[width=40mm]{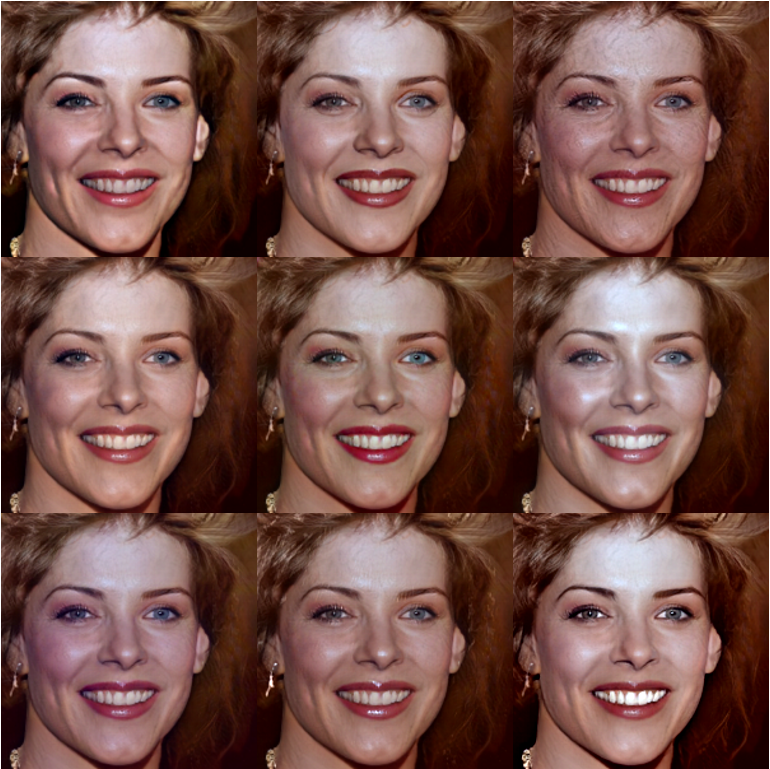}    
  \caption{Decomposition of faces. Top: the output of the 9 branches  (with increased contrast). The sum of all branches produce the generated images, Fig. \ref{fig:gan_generated}. Bottom: each branch output (channel) is added (without amplification) to the generated image, to visualize its role.
  We clearly observe each branch is specializing in different image characteristics.  }
  \label{fig:gan_decomp}
\end{figure}
\begin{figure}[htb]
  \centering
  \includegraphics[width=40mm]{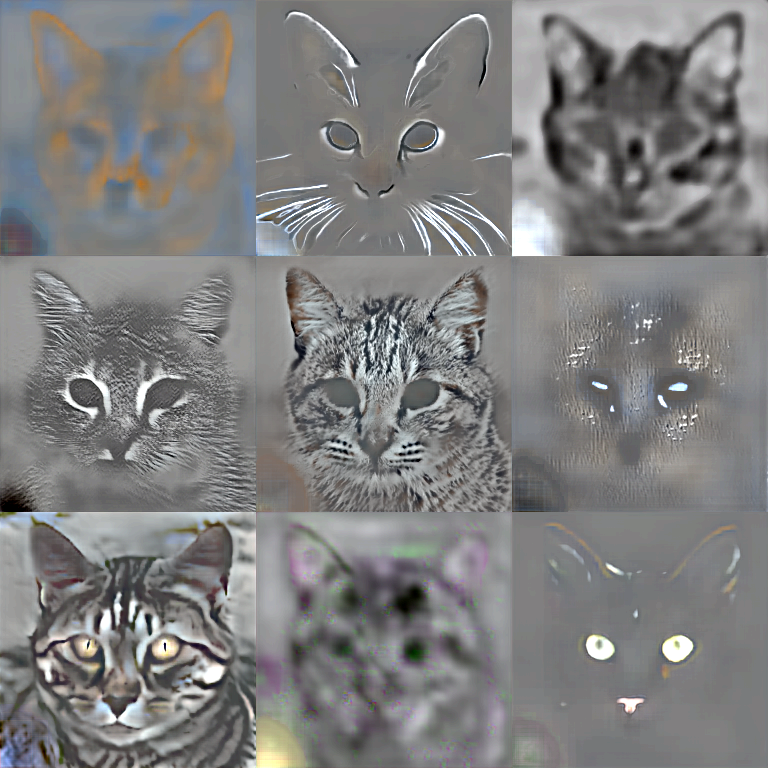}
  \includegraphics[width=40mm]{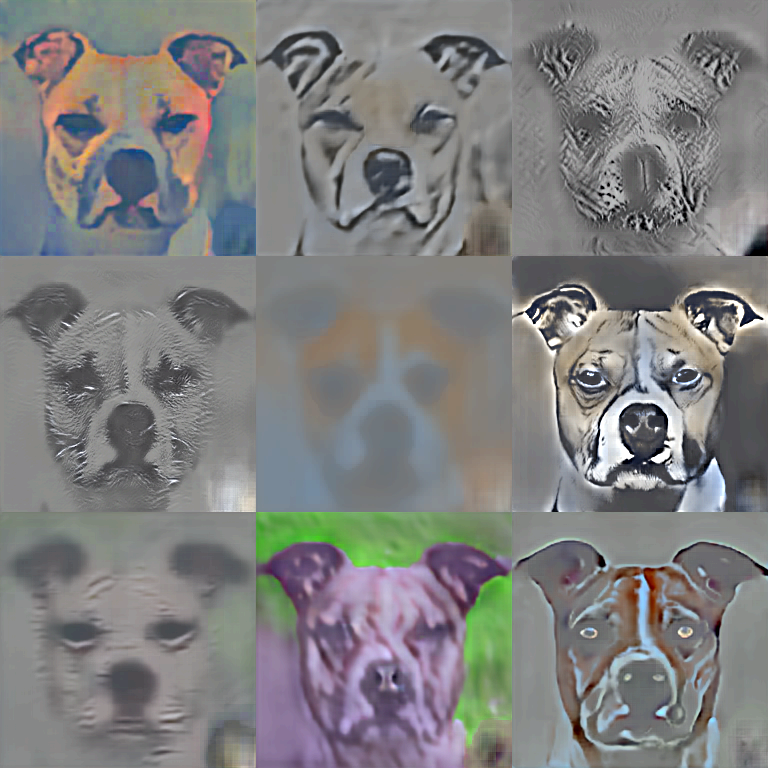}
  \includegraphics[width=40mm]{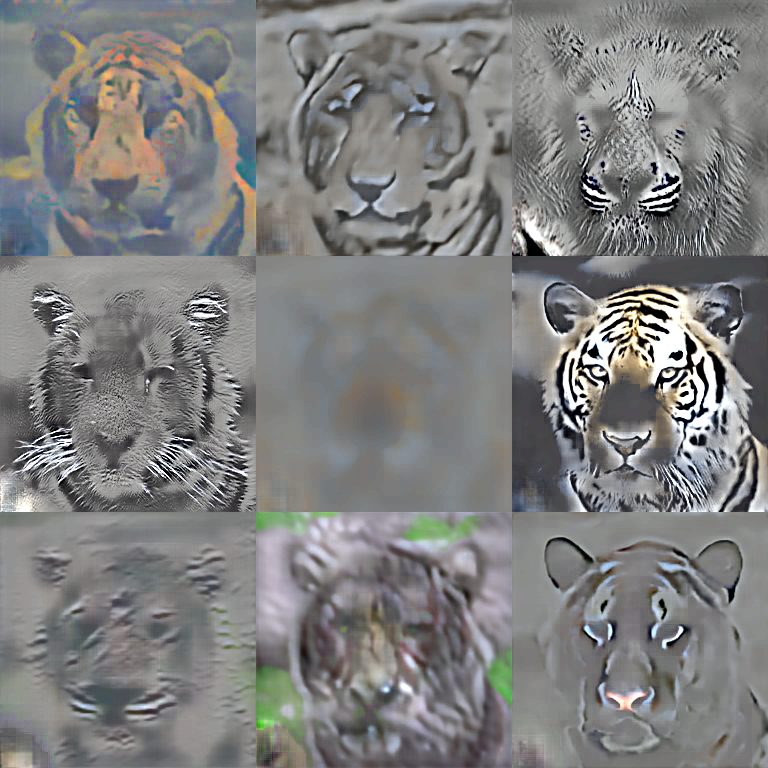}
  \caption{Decomposition of animal images as outputs of the 9 branches  (with increased contrast). The sum of all branches produce the generated images in Fig. \ref{fig:gan_generated}, excluding $R$. }
  \label{fig:gan_decomp_animals}
\end{figure}


We train our NN as a GAN, but do not use any of the additional loss terms originally used by  StarGAN-v2. Cycle loss is discarded as well, and fidelity to the input image is attained by slightly modifying  Eq. \eqref{eq:f} as follows:  $f_{\theta}(x) = R + \sum_{k=1}^M v_k(x) $, where $R$ is an over-smoothed version of $x$.

We added an optional learnt nonlinear diffusion step, which produces $R$ and pre- processes the inputs. This results in higher quality decompositions, nevertheless - image decomposition occurs also without this step, as shown in the supplementary material, where full details of the architecture and loss are provided as well.

As in previous experiments, different branches specialize w.r.t. a different aspect of the task at hand - in this case it is generating natural images. By construction - $v_k(x)$ are ambient-space representations, summed to generate the image - hence we say that the branch outputs are "decompositions of the image". Remarkably, these specializations are mostly interpretable: Celeb-a-HQ images seem to decompose to  specular light reflection, diffusive light reflection, color hue, and texture. AFHQ images are decomposed to color patterns, different fur textures, whiskers and contours, color hue, eyes and nose. See Fig. \ref{fig:gan_generated} for re-generated images, and Fig. \ref{fig:gan_decomp} for their learnt specialized decompositions.

These decompositions may be useful for a variety of tasks. As an example, we devised a segmentation procedure for cat's eyes, see demonstration in Fig. \ref{fig: segmented cats}, and Full details in the supplementary.
Further applications can include 3D shape recovery from lighting, random channel perturbations for data augmentation, image manipulation and filtering and more (see the supplementary for some examples).

\begin{figure}[htb]
  \centering
  \includegraphics[width=0.53\textwidth]{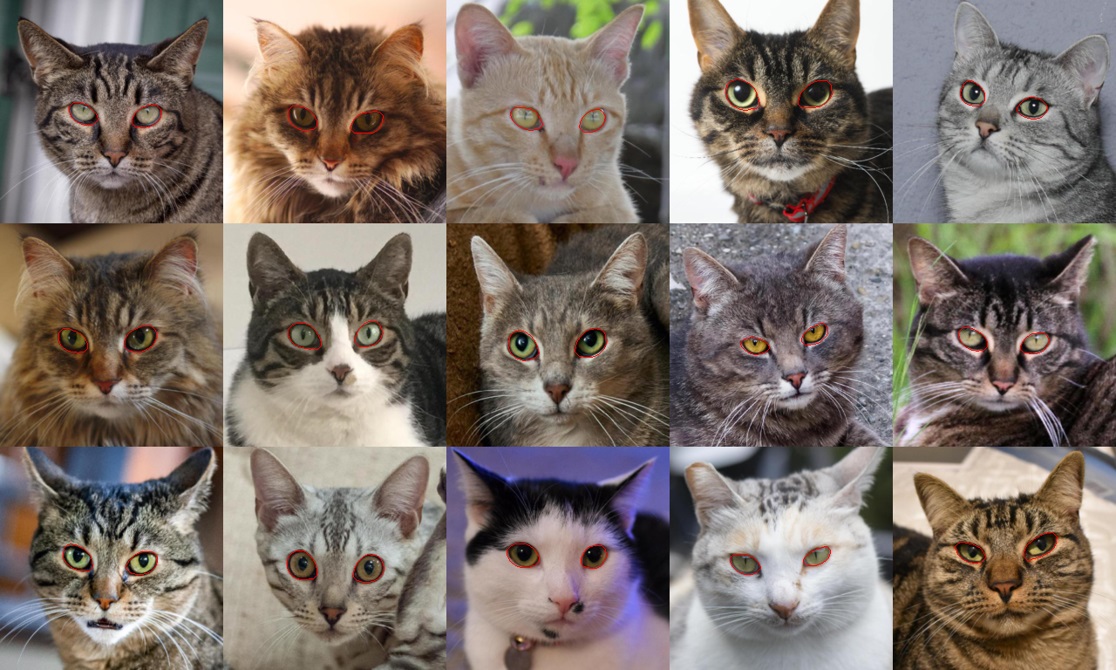}
  \includegraphics[width=0.081\textwidth]{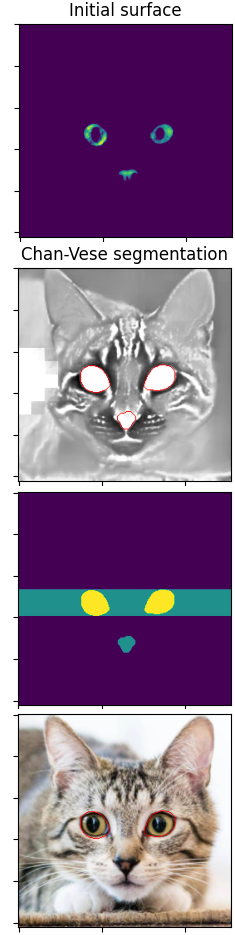}
  \caption{Segmentation of cat's eyes. Only class labels (dog/ cat/ wild)  are used with no segmentation labels. Right: Decompositions are used for inputs of Chan-Vese segmentation algorithm \cite{chan2001active}.}
  \label{fig: segmented cats}
\end{figure}



\section{Discussion of Branch Equilibrium}
Here we extend Sec. \ref{sec:mathematical}. A branch $v_k$ attains equilibrium when $\dot{\theta}_k=0$. In SGD this happens when $\nabla_{\theta_k} L(x)=0,\, \forall x \in X$. Denote $D_{collab}(x)=\frac{d L}{d f_{\theta}(x)} \in \mathbb{R}^C$ and $D_{dist}^k(x)= \nabla_{\theta_k} v_k(x) \in \mathbb{R}^{C \times |\theta_k|}$.

Consider $D_{collab}(x)=0$. We make the sensible assumption that this happens for a small subset of $X$. Explanation - plugging Eq. \eqref{eq:cross_entropy_gradient} or \eqref{eq:collaborative_L2} to $D_{collab}(x)=0$ translates to a perfect fit by the model. Usually in machine learning, a successfully trained model does not fit perfectly most of of the data.

Consider $\nabla_{\theta_k} L(x)=0,\,D_{collab}(x) \neq 0$. Then by Eq. \eqref{eq:collab_distrib}, $D_{dist}^k(x) $'s columns lie in a $C-1$ dimensional plane, orthogonal to $D_{collab}$. Because  $|\theta_k| \gg C$, we assume orthogonality is attained for a small subset of the columns, and the rest are zero columns. Otherwise we get high linear dependency between the columns in the near neighbourhood of equilibrium, which is usually unlikely.


Re-phrasing the above, when $\nabla_{\theta_k} L(x)=0$ is attained, Eq. \eqref{eq:collab_distrib, scalar} has a  zero distributive part for most of $(w_k, x)$ pairs. Under this mechanism different branches may reach equilibrium independently. There are many open problems, such as when is linear dependency of the gradients expected? How much data do we expect to fit perfectly?

To conclude, gradients of a branched model are split to distributive and collaborative factors \eqref{eq:collab_distrib}. Inter-branch Hessian entries have the same distributive factor \eqref{eq:mixed_jacobian_entries}. We conjecture the distributive part is the main factor in driving stability. In such a case - the Hessian is block diagonal, where blocks describe branches, i.e. perturbations in one branch do not cause loss gradients to change any of the other branches. Thus each branch minimizes the loss function from an independent aspect.




\section{Conclusion}

Branch Specialization is observed in all experiments as different branches naturally solve different sub-tasks of the global task at hand. In the context of classification, we show that different branches specialize at classifying different types of data samples of the same class (such as close horses in one branch and far-away horses in another).  In the context of generative models, we found that specialization is manifested as a first of its kind image decomposition. We envision these decompositions can be harnessed for self-supervised learning - as we demonstrate for segmentation.
Preliminary analysis of gradient descent on branched architectures has shown that (under sensible assumptions) there is a strong inclination towards independent local minima of each branch, which do not interfere with the optimization process of other branches. 


\bibliographystyle{plainnat}  

\bibliography{references.bib}

\newpage
\appendix

\section{Class Transfer GAN: Ambient Space Image Manipulation}

\begin{figure}[!htb]
  \centering
  \includegraphics[width=\textwidth]{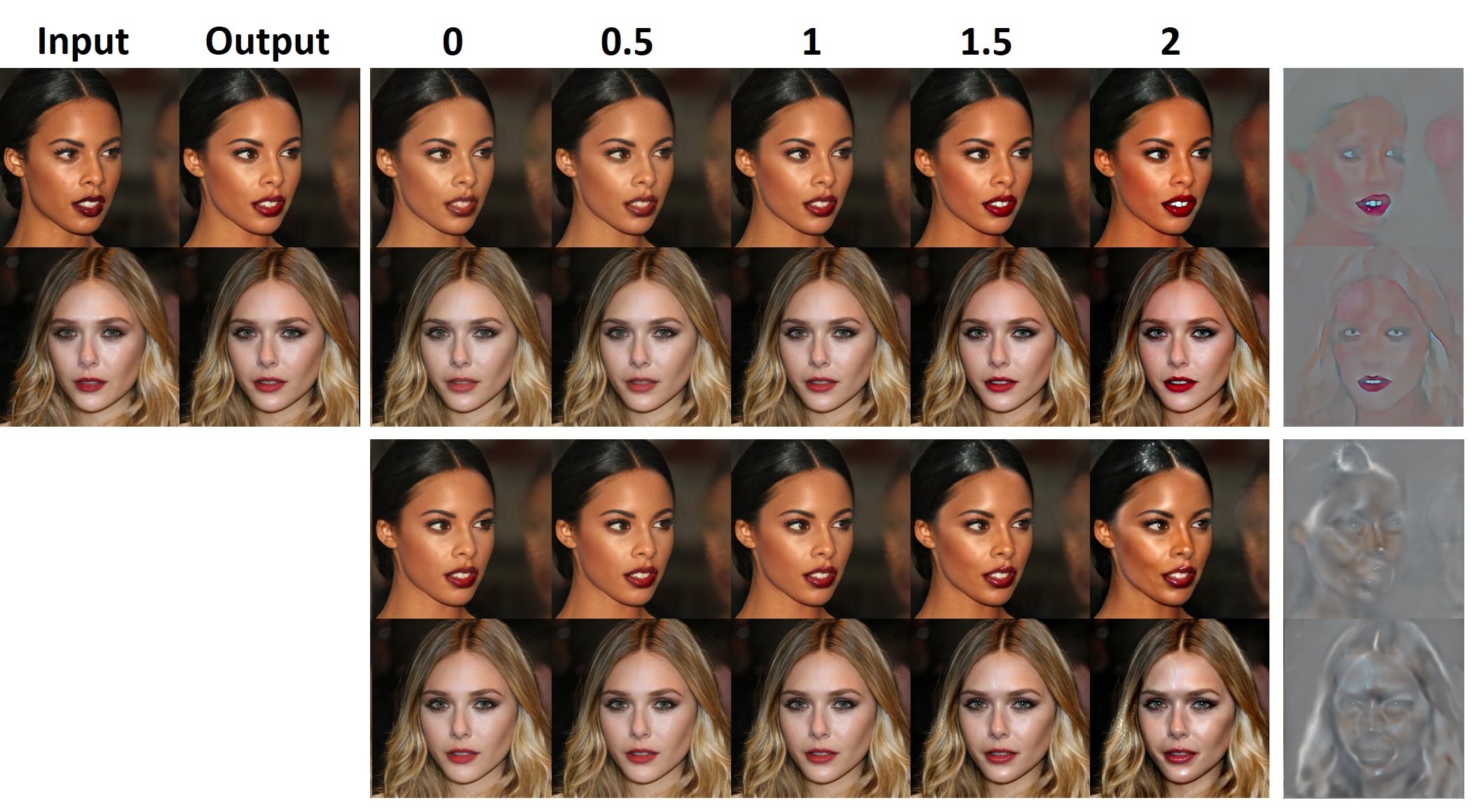}
  \caption{Top left: Images and their re-generation; Right: Two ambient-space components, of make-up (top) and lighting (bottom); Middle: Sweeping magnitude of each ambient-space component in the re-generated image.}
  \end{figure}
  
  \begin{figure}[!htb]
  \centering
  \includegraphics[width=\textwidth]{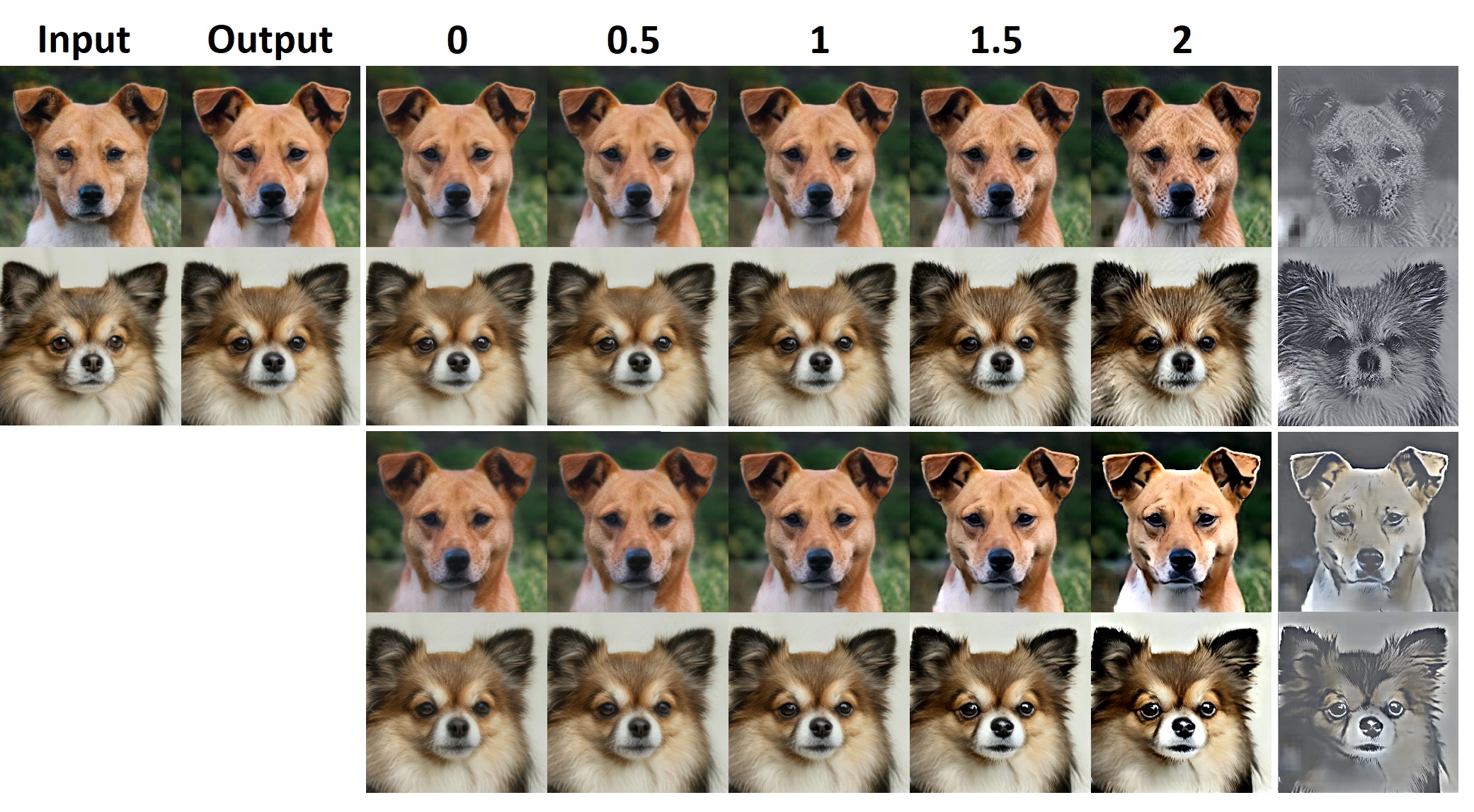}
  \caption{Top left: Images and their re-generation; Right: Two ambient-space components, of fur (top) and comic-like-celluloid component (bottom); Middle: Sweeping magnitude of each ambient-space component in the re-generated image.}
  \end{figure}

\begin{figure}[!htbp]
  \centering
  \includegraphics[width=\textwidth]{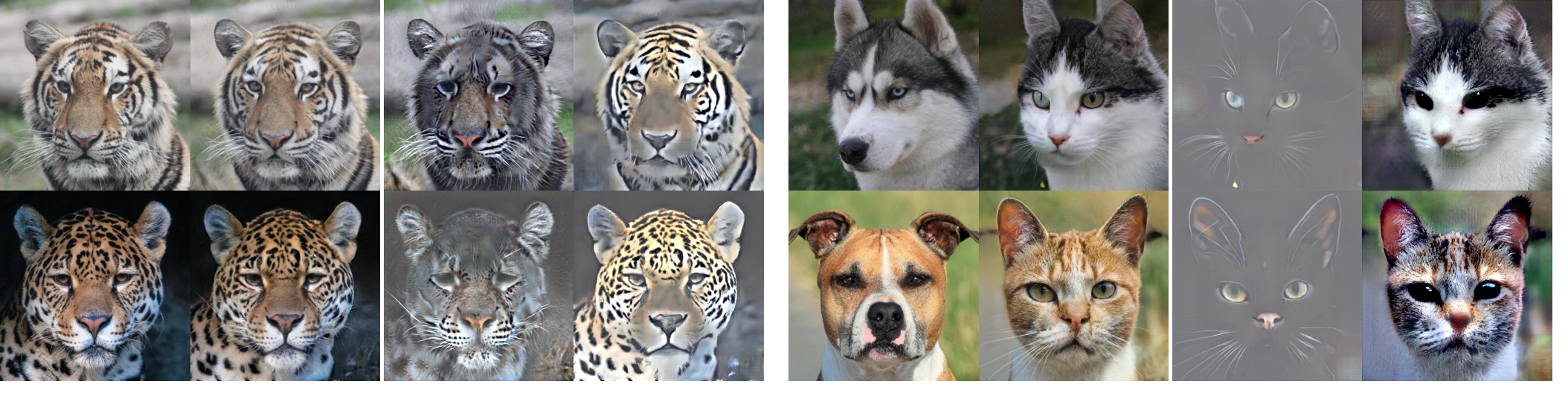}
  \caption{Classes Cat and Wild: Images, their re-generation, and separation between two types of textures by summing different subsets of the ambient-space components.}
  \end{figure}

\newpage
 \section{Cifar10 Classifiers Intra-Branch Specialization}

\begin{figure}[!htbp]
  \centering
  \includegraphics[width=\textwidth]{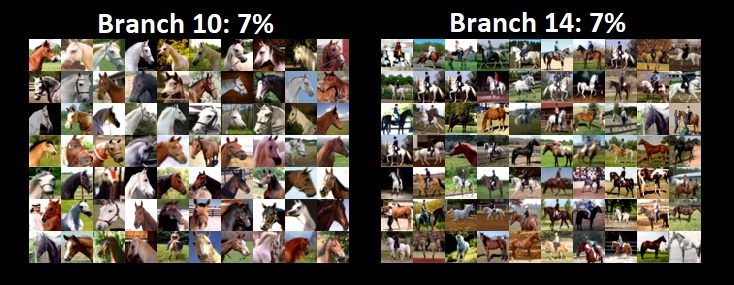}
  \caption{Branches 10, 14: We examine horse samples of top $7\%$ confidence, where confidence was assessed for each branch separately - as described in Fig. 8 in the main article. Confidence values of these branches continue to be higher for either close-ups or full body shots of horses even at $7\%$.}
 \end{figure}

\newpage
\section{Cifar10 Classifiers with various No. of Branches}

\begin{figure}[!htbp]
  \centering
  \includegraphics[width=0.65\textwidth]{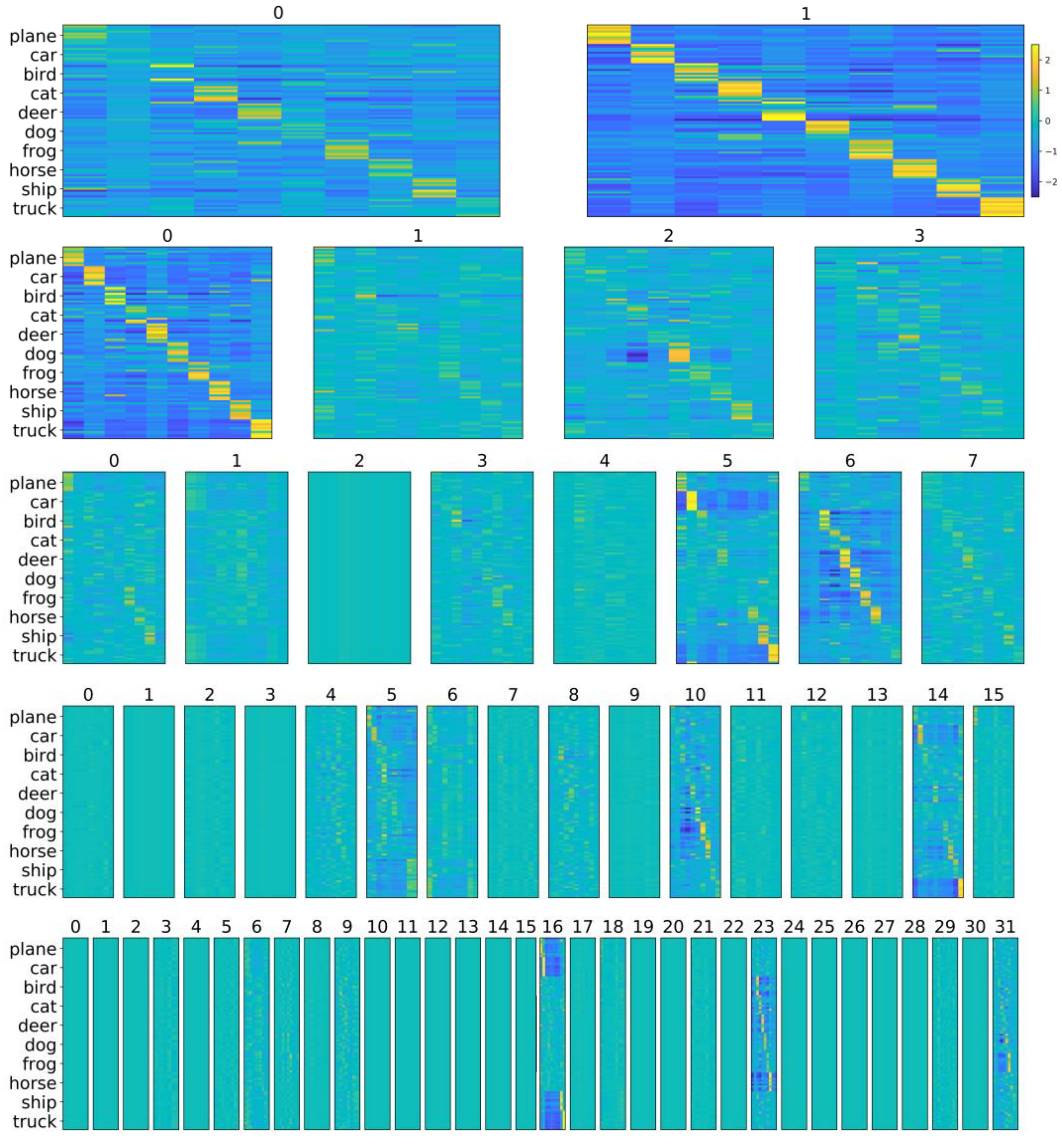}
  \includegraphics[width=0.2\textwidth]{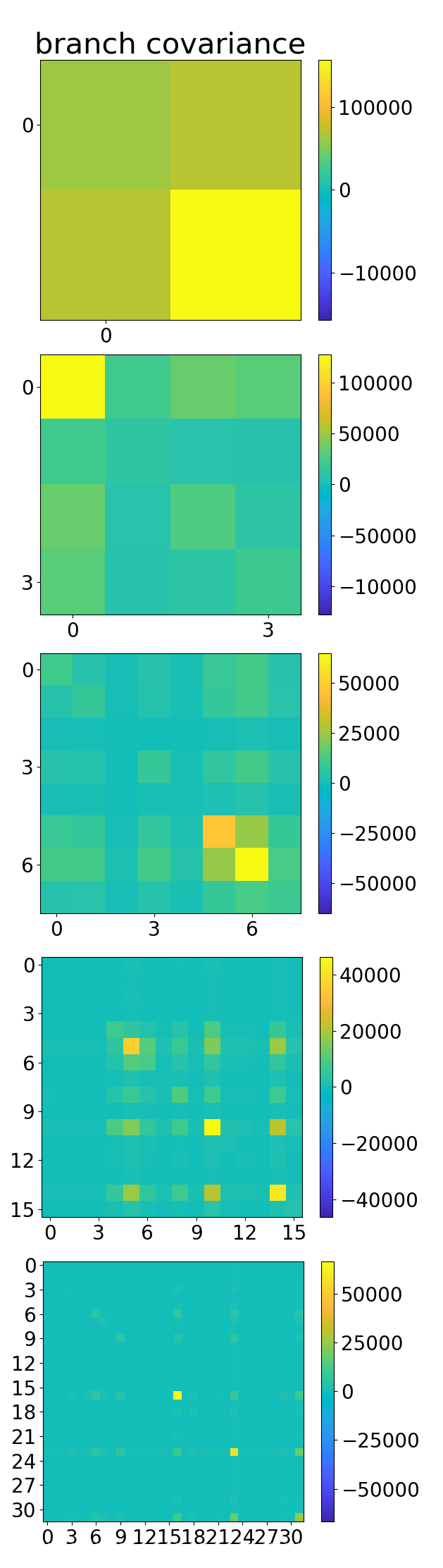}
  \caption{Left: Branch outputs for models with $M=[2, 4, 8, 16, 32]$ branches - as seen in Fig. 6 in the main paper. The No. of Silent and Active Branches, as quantified in Fig. 5 in the main paper, is calculated according to the branch outputs shown here.
  Right: Covariance matrices, as in Fig. 5 in the main paper}.
  \end{figure}

\section{Class Transfer GAN: with and without learnt pre-processing}
\begin{figure}
  \includegraphics[width=0.9\textwidth]
  {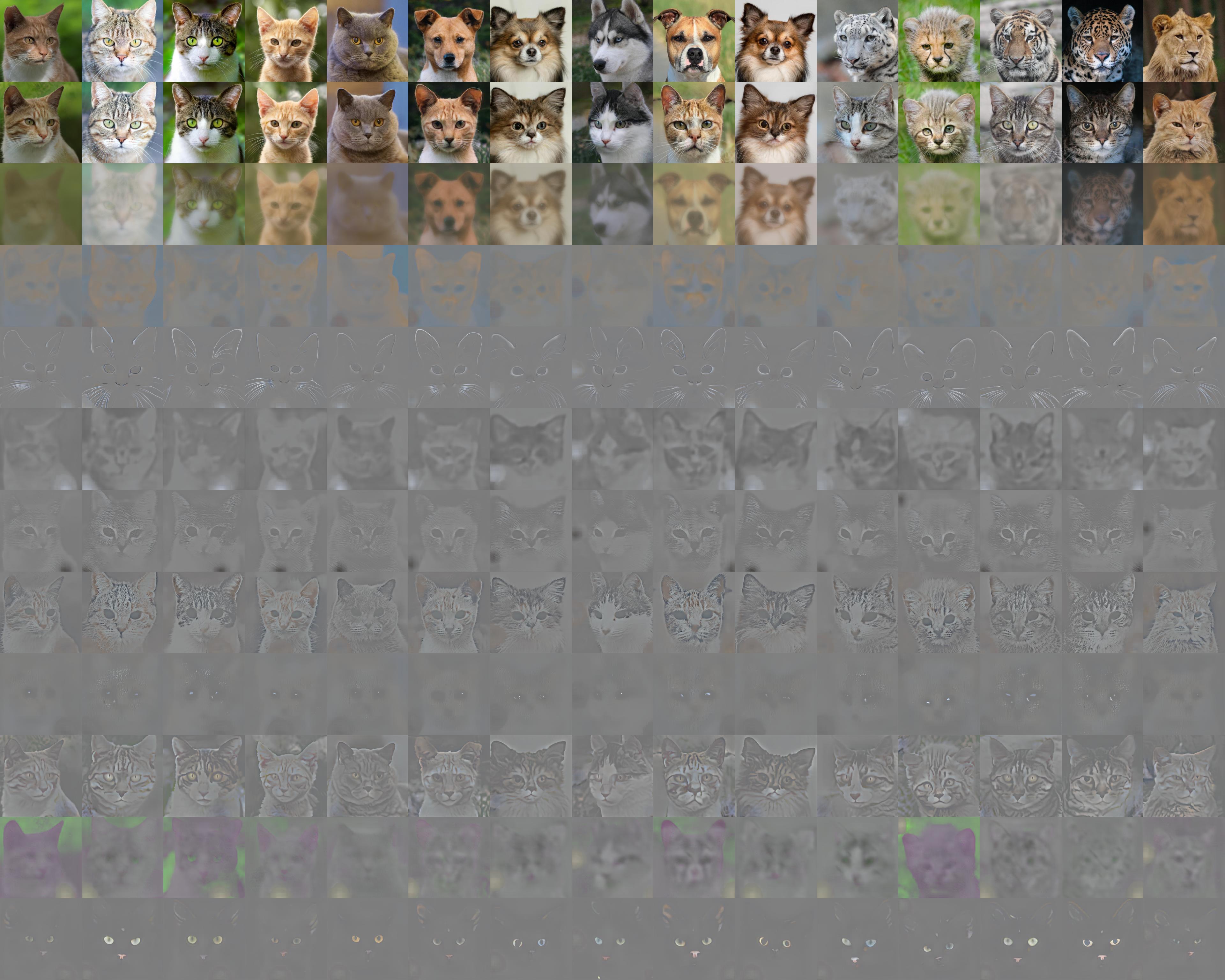}
  \includegraphics[width=0.9\textwidth]
  {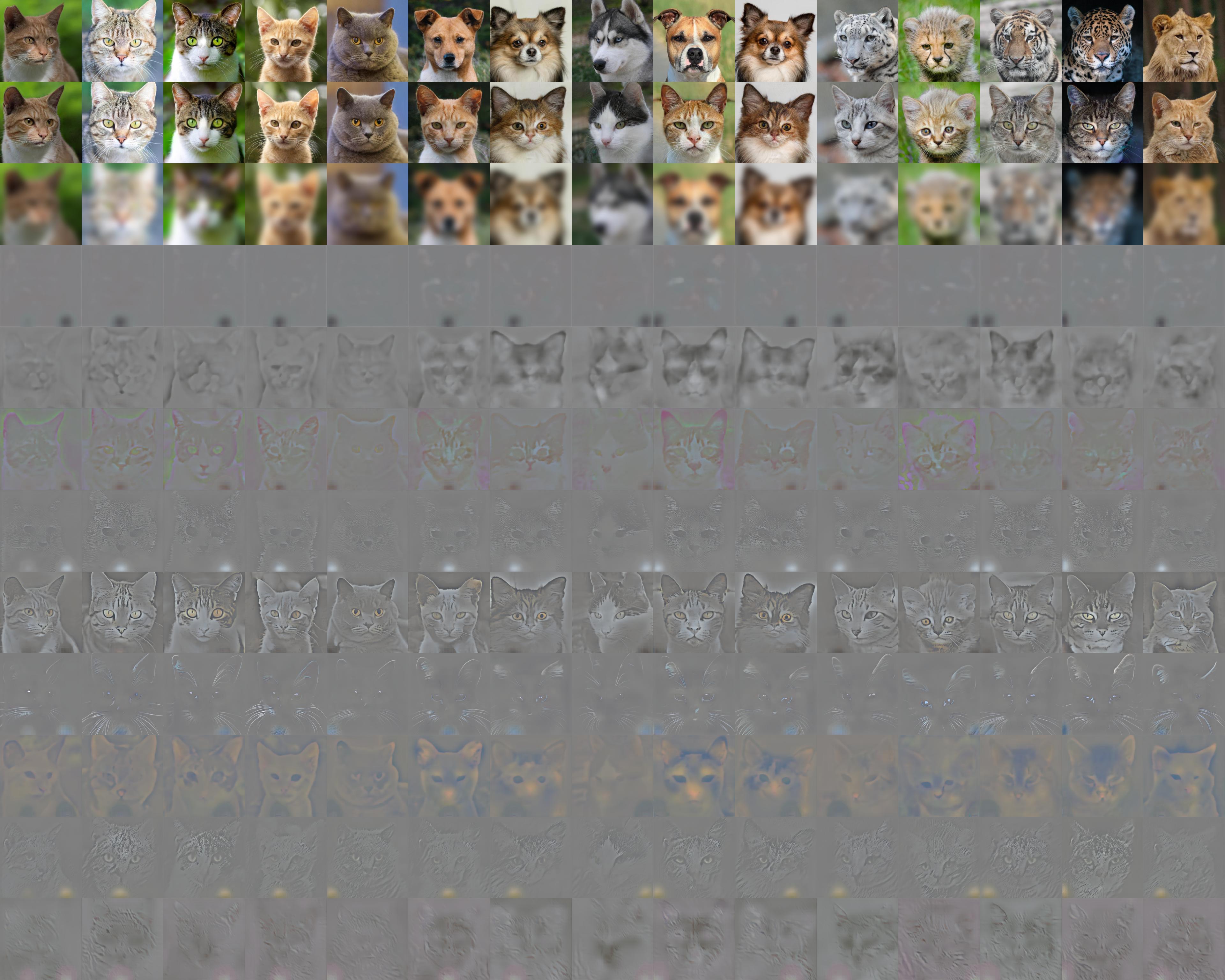}
  \caption{Cats: Images, their re-generation, their $R$ and outputs of the 9 branches of the generator network. Top: With learnt non-linear diffusion pre-processing,  each branch outputs an ambient-space representation, where branches differ by textures, colors, and segments; Bottom: Without non-linear diffusion pre-processing, i.e. each branch recieves $x$ as a direct input and $R$ is a Gaussian-smoothed image. Similar ambient-space decompositions are obtained, but quality is inferior.} \label{fig:cats_decomp}
\end{figure}

\begin{figure}
  \includegraphics[width=0.9\textwidth]
{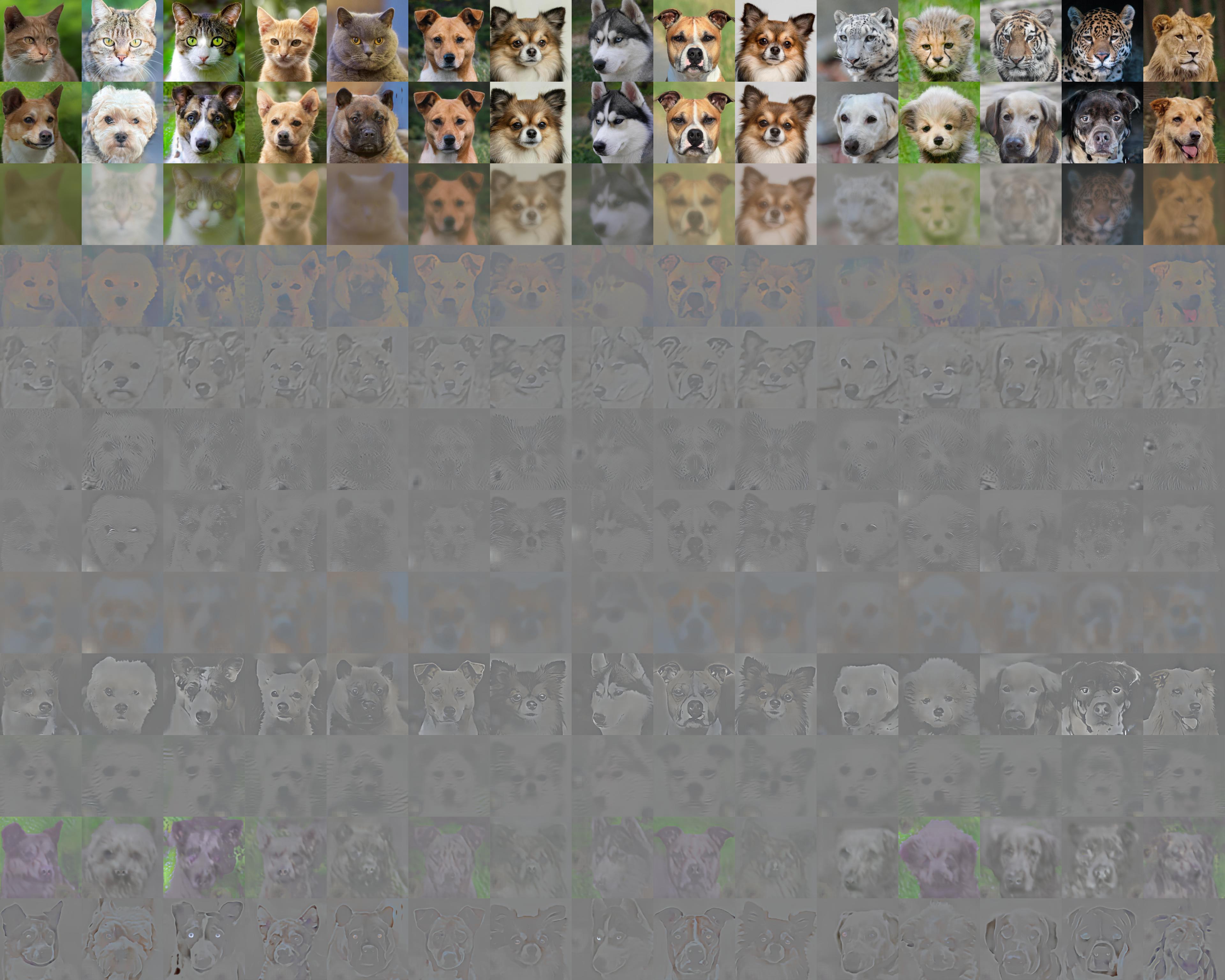}
 \includegraphics[width=0.9\textwidth]
  {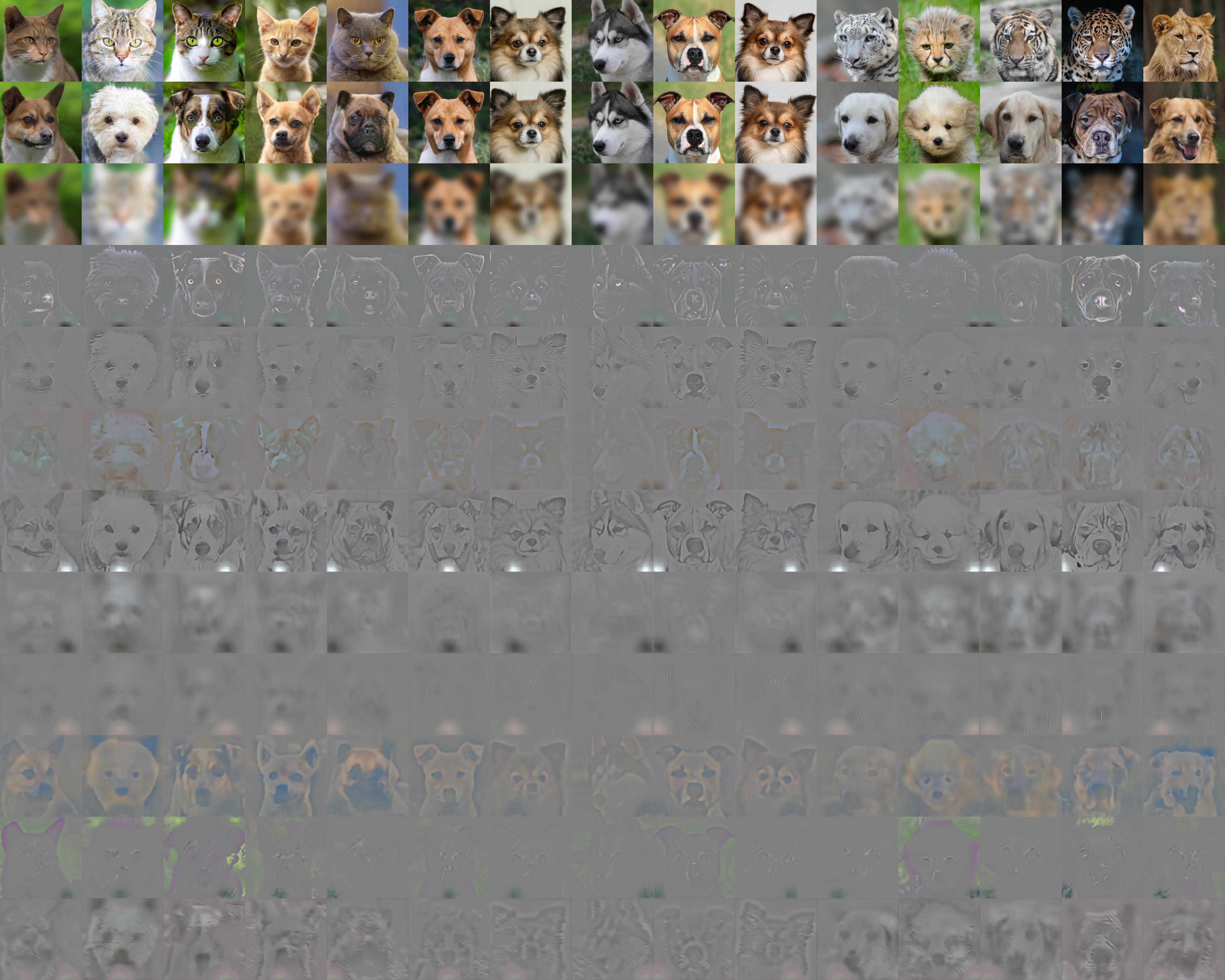}
  \caption{Dogs: Similarly to Fig. \ref{fig:cats_decomp}.}\label{fig:dogs_decomp}
\end{figure}

\begin{figure} 
  \includegraphics[width=0.9\textwidth]
{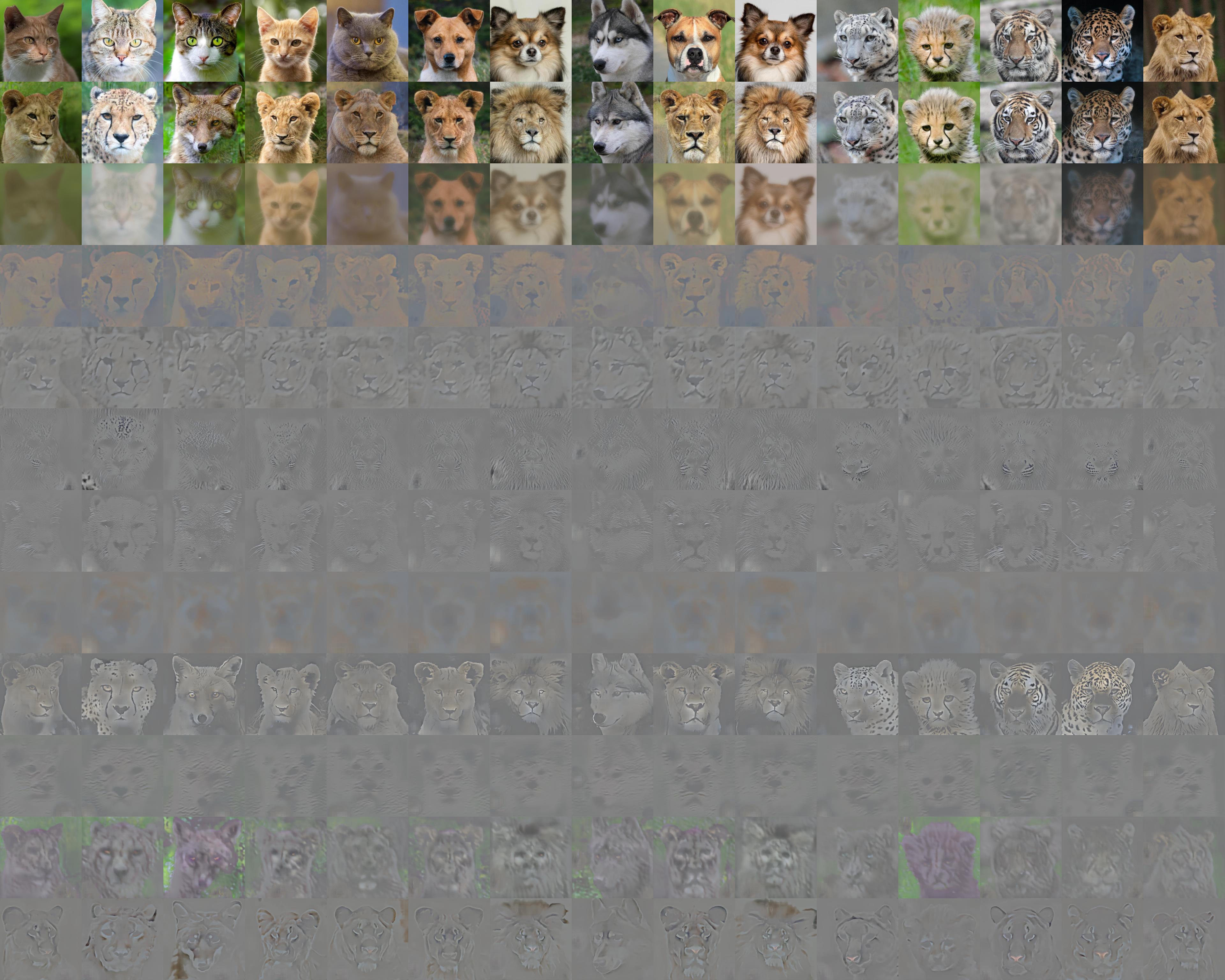}
 \includegraphics[width=0.9\textwidth]
  {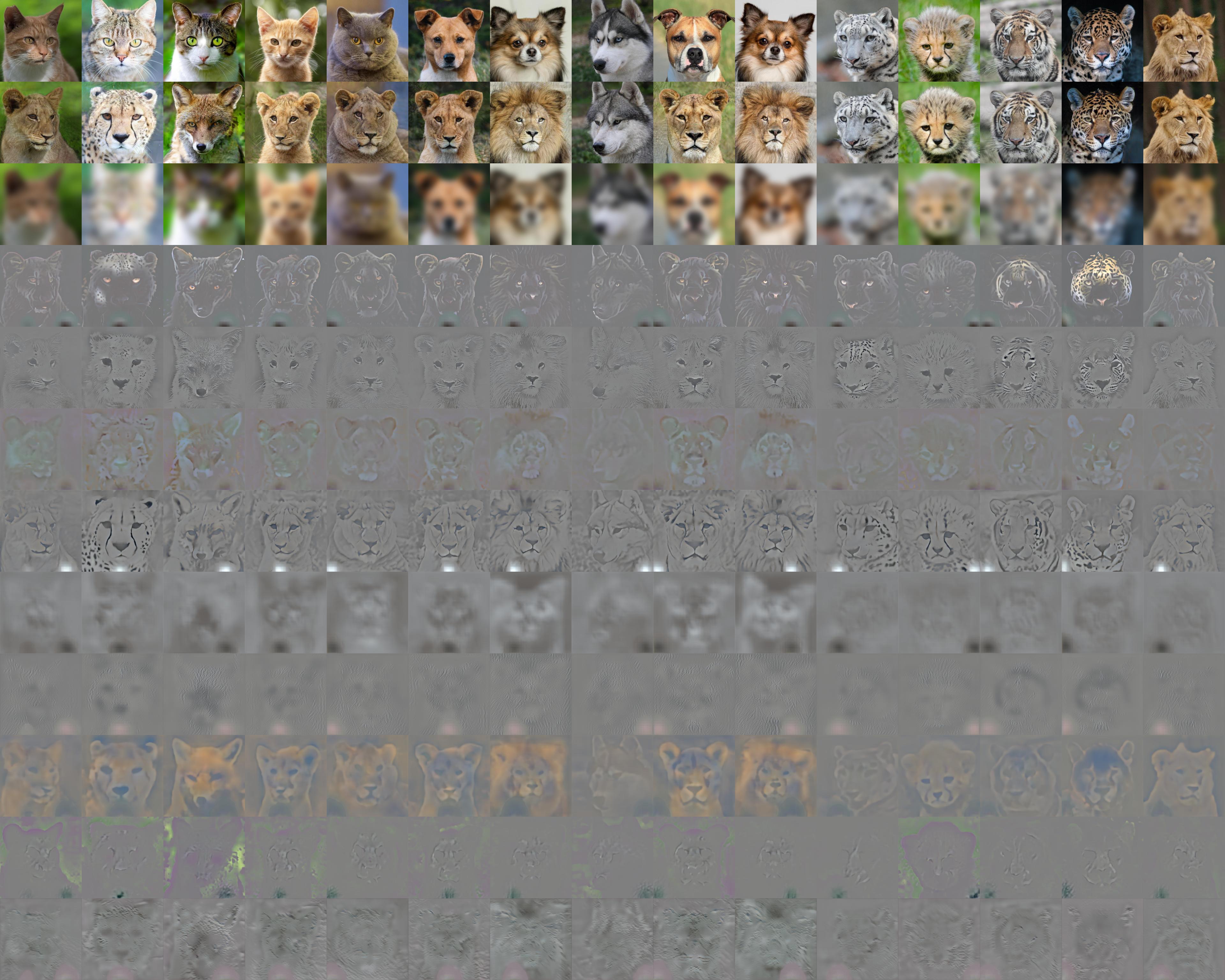}
  \caption{Wild: Similarly to Fig. \ref{fig:cats_decomp}.}\label{fig:wild_decomp}
\end{figure}

In the main paper we add a nonlinear diffusion pre-processing of the input (details in Sec. \ref{sec:stargan_details}). Here we show that branch outputs of the Class Transfer GAN are ambient-space representations with this addition, and without it as well (Figs.  \ref{fig:cats_decomp}, \ref{fig:dogs_decomp}, \ref{fig:wild_decomp}).

\newpage
\section{Class transfer GAN: Loss and optional Nonlinear Diffusion of the input}
 \label{sec:stargan_details}To have a conciser meaning of the training objective, we train our NN as a GAN, but do not use any of the additional loss terms originally used by StarGAN-v2. This requires two adaptations: First, the style diversification and style reconstruction losses are simply discarded, as we do not require diverse synthesis. Second the cycle-loss is discarded, and fidelity to input image is attained by slightly modifying  Eq. (1) of the main paper as follows: \begin{equation} \label{eq:klavtul_no_spectral}
f_{\theta}(x) = R + \sum_{k=1}^M v_k(x) ,
\end{equation}
where $R$ is an over-smoothed version of $x$ - either learnt smoothing or Gaussian smoothing.

Additionally - we added a learnt pre-processing step, which learns $R$, and processes inputs: This procedure was devised as a neural-network that suffices $R$, as well as lowly-correlated inputs for each branch. The nonlinear diffusion is implemented as a recurrent neural-network with architecture $p(x; W)$, hence the nonlinear diffusion process is
\begin{equation}
u_t(t)=-p_{}(t),\,u(0)=x \in X,
\end{equation}
where time steps are often implemented discretely. This was done before, for instance \cite{chen2016trainable}. Let $\phi(t) = t \cdot u_{tt}$ then we can compute a reconstruction formula, for a general stopping time $T$, using integration by parts (and assuming $u_t(0)$ is bounded)

\begin{equation}\label{eq:rec}
\int_0^T \phi(t)\,dt = t u_t|_0^T - \int_0^T u_t\,dt =T u_t(T)-u(T)+x=-Tp(u(T))-u(T)+x.
\end{equation} 
In other words - let  $R = T p(u((T)) + u(T)$, then the following reconstruction identity holds $x=\int_0^T \phi(t)\,dt + R$. This holds for a discretized nonlinear diffusion process as well, where integrals are replaced by weighted  sums, and time derivatives by discrete time derivatives. This is a similar mechanism to Spectral TV \cite{gilboa2014total}.

Because $R$ is the residual of a nonlinear diffusion process, driven by a CNN, it generalizes the regular Gaussian smoothing. Hence we use it as a learnt over-smoothing of the image. $\phi(t)$ capture different parts of the nonlinear diffusion process - hence for a smoothing process, it suffices a multi-scale decomposition of $x$. $\phi$ alongside $R$ enables full reconstruction of the input $x$ \eqref{eq:rec}. Discretizing the diffusion process to $M+1$ steps, we a obtain a discrete set $\{\phi_k\}_{k=1}^M,\,\phi_i \in X$, and each $\phi_k$ is fed to a different branch.

Thus the inference becomes $f_{\theta}(x) = R + \sum_{k=1}^M v_k(\phi_k) $.  The architecture of $p$ is an 8-layer bottleneck CNN with $4.05 \cdot 10^4$ parameters, where input and output dimensions are the same and equivalent to dimension of $x$.  Evidently (Figs. \ref{fig:cats_decomp}, \ref{fig:dogs_decomp}, \ref{fig:wild_decomp}), $p$ is learnt to be a smoothing operator - as $R$ is indeed a smoothed version of $x$.

Remark: We noticed that throughout the training procedure $R$ may fluctuate between a slightly smoothed version  $x$ to a significantly smoothed $x$ - at times, even a constant constant image.

\newpage
\section{Unsupervised Segmentation }
\begin{figure}[!htbp]
  \centering
  \includegraphics[width=0.5\textwidth]{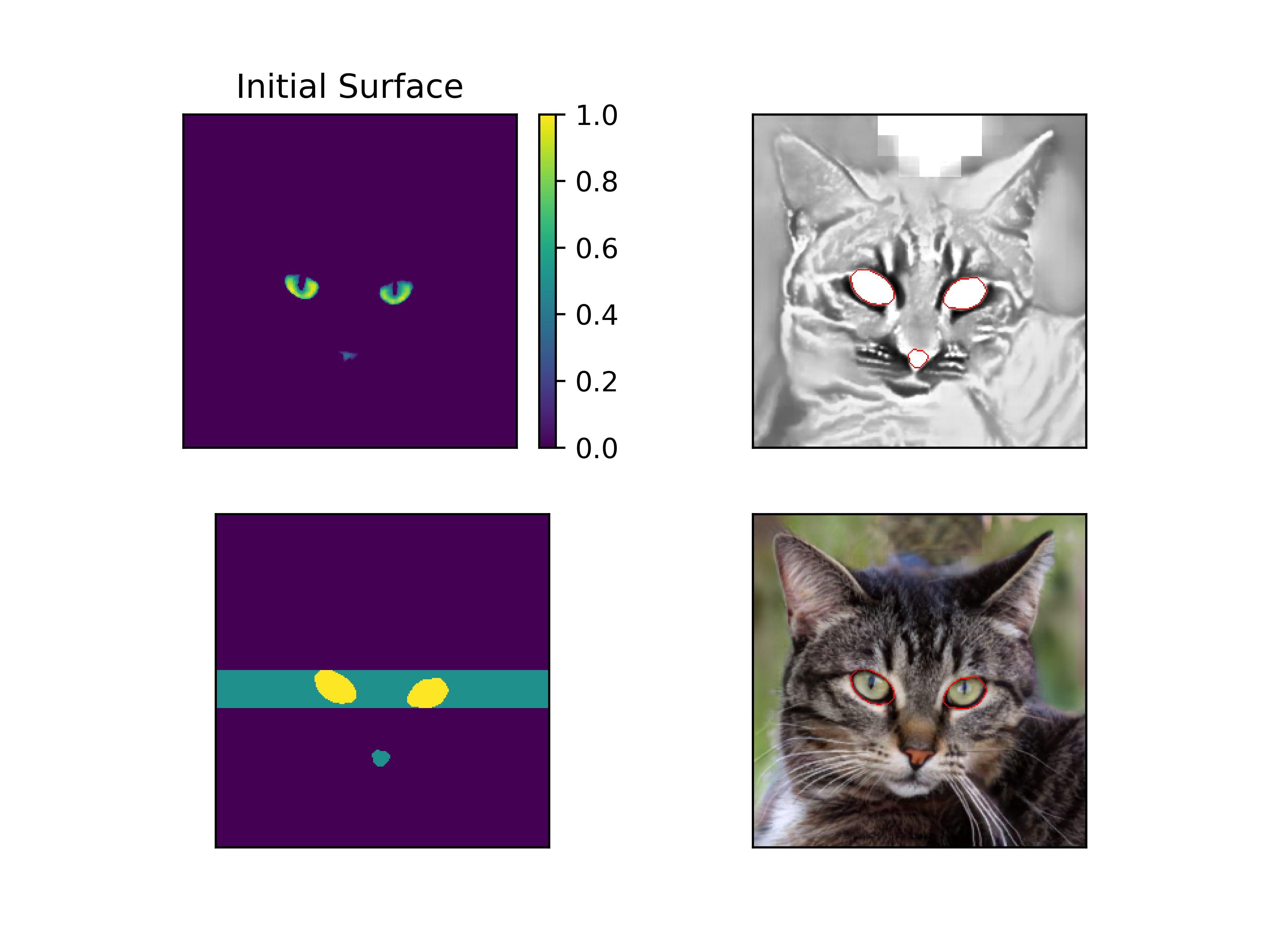}
  \caption{In the main paper, we demonstrate how our decompositions, which are ambient-space components, may be useful for segmentation. Because the components are in ambient-space, they are completely aligned with the generated image, and segmentation can be done directly upon them. We segment cats' eyes. We use the Chan-Vese \cite{chan2001active} algorithm to optimize a surface-image so that image-regions, which correspond to the positive regions of the surface-image, have low variance. Some of our components are good at locating eyes and nose - thus they are used for the initial surface. Other components tend to have constant values in the eye regions - hence we use these as the image we segment. Finally - this procedure often captures the nose as well - so as a final criterion, we filter the connected components that have row-wise multiplicity of 2. Remark: A pre-processing step entails converting the RGB components to scalar images. We found that using the Value channel of HSV works nicely.}
  \end{figure}


\section{Computational Resources}
We use VGA compatible controller: NVIDIA Corporation GV100GL [Tesla V100 DGXS 32GB] (rev a1).


\end{document}